\documentclass[10pt,twocolumn,letterpaper]{article}

\usepackage[pagenumbers]{cvpr} %
\usepackage{graphicx}
\usepackage{multicol}
\usepackage{tabularx}
\usepackage{amsmath}
\usepackage{amssymb}
\usepackage{float}
\usepackage{caption}
\usepackage{array}
\usepackage{xcolor}
\usepackage{microtype}

\makeatletter
\renewcommand{\paragraph}{%
  \@startsection{paragraph}{4}%
  {\z@}{0.5ex \@plus 0.5ex \@minus .2ex}{-1em}%
  {\normalfont\normalsize\bfseries}%
}
\makeatother
\usepackage{arydshln}
\usepackage{microtype}
\usepackage[T1]{fontenc}
\usepackage{chngcntr}
\usepackage[utf8]{inputenc}

\definecolor{cvprblue}{rgb}{0.21,0.49,0.74}
\usepackage[pagebackref,breaklinks,colorlinks,citecolor=cvprblue]{hyperref}

\def\paperID{5198} %
\def\confName{CVPR}
\def\confYear{2024}

\title{Concept Sliders: LoRA Adaptors for Precise Control in Diffusion Models}

\author{Rohit Gandikota$^{1}$ \hspace{1.5em} Joanna Materzy\'nska$^{2}$ \hspace{1.5em} Tingrui Zhou$^{3}$\hspace{1.5em} Antonio Torralba$^{2}$ \hspace{1.5em} David Bau$^{1}$ \vspace{3pt} \\ 
$^{1}$Northeastern University  \quad $^{2}$Massachusetts Institute of Technology \quad $^{3}$ Independent Researcher }

\begin{document}
\twocolumn [{
\renewcommand\twocolumn[1][]{#1}%
\maketitle
\includegraphics[width=\textwidth]{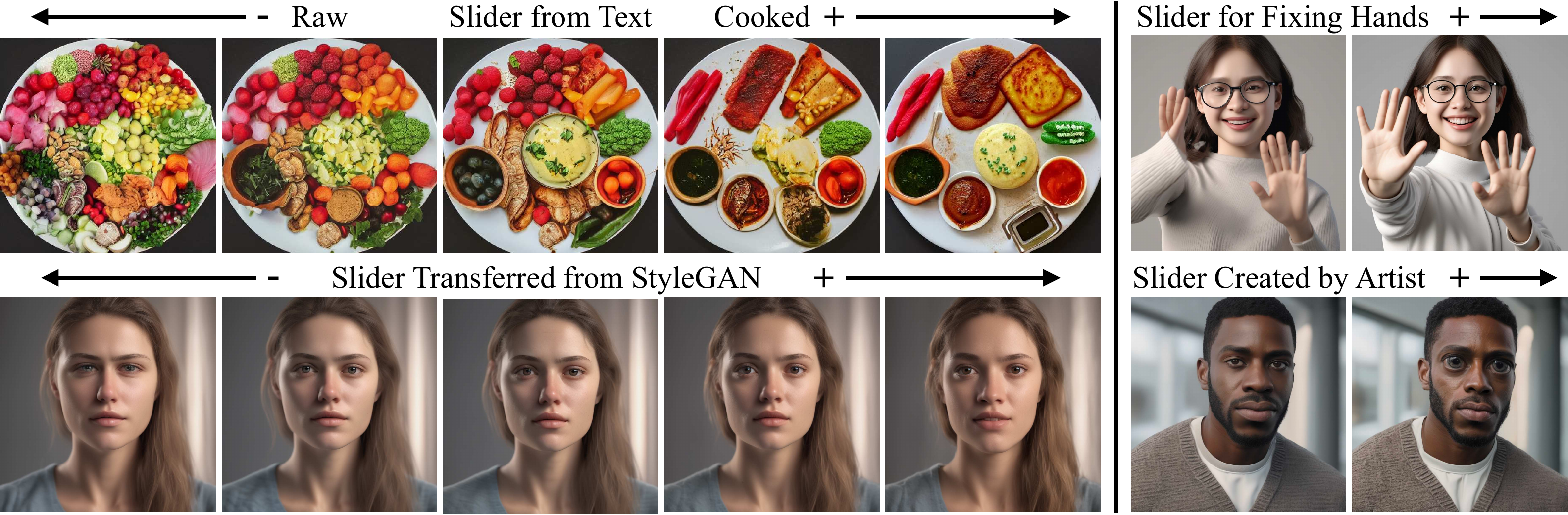}
\captionof{figure}{Given a small set of text prompts or paired image data, our method identifies low-rank directions in diffusion parameter space for targeted concept control with minimal interference to other attributes. These directions can be derived from pairs of opposing textual concepts or artist-created images, and they are composable for complex multi-attribute control. We demonstrate the effectivness of our method by fixing distorted hands in Stable Diffusion outputs and transferring disentangled StyleGAN latents into diffusion models. \vspace{2em}}
\label{fig:intro}
}]

 \begin{abstract}
\vspace{-5pt}We present a method to create interpretable concept sliders that enable precise control over attributes in image generations from diffusion models. Our approach identifies a low-rank parameter direction corresponding to one concept while minimizing interference with other attributes. A slider is created using a small set of prompts or sample images; thus slider directions can be created for either textual or visual concepts. Concept Sliders are plug-and-play: they can be composed efficiently and continuously modulated, enabling precise control over image generation. In quantitative experiments comparing to previous editing techniques, our sliders exhibit stronger targeted edits with lower interference. We showcase sliders for weather, age, styles, and expressions, as well as slider compositions. We show how sliders can transfer latents from StyleGAN for intuitive editing of visual concepts for which textual description is difficult. We also find that our method can help address persistent quality issues in Stable Diffusion XL including repair of object deformations and fixing distorted hands. Our code, data, and trained sliders are available at \href{https://sliders.baulab.info/}{sliders.baulab.info}
\end{abstract}

 \section{Introduction}
Artistic users of text-to-image diffusion models~\cite{dhariwal2021diffusion,rombach2022high,podell2023sdxl,imagen2022webpage,betker2023improving} often need finer control over the visual attributes and concepts expressed in a generated image than currently possible. Using only text prompts, it can be challenging to precisely modulate continuous attributes such as a person's age or the intensity of the weather, and this limitation hinders creators' ability to adjust images to match their vision~\cite{slider_reddit}. In this paper, we address these needs by introducing interpretable \emph{Concept Sliders} that allow nuanced editing of concepts within diffusion models. Our method empowers creators with high-fidelity control over the generative process as well as image editing. Our code and trained sliders will be open sourced. \par

\let\thefootnote\relax\footnotetext{$^{1}${\tt\small [gandikota.ro,davidbau]@northeastern.edu} $^{2}${\tt\small [jomat, torralba]@mit.edu}\\ $^{3}${\tt\small shu\_teiei@outlook.jp}}

Concept Sliders solve several problems that are not well-addressed by previous methods.  Direct prompt modification can control many image attributes, but changing the prompt often drastically alters overall image structure due to the sensitivity of outputs to the prompt-seed combination~\cite{ruiz2022dreambooth, wu2023uncovering, kawar2022imagic}. Post-hoc techniques such PromptToPrompt~\cite{hertz2022prompt} and Pix2Video~\cite{ceylan2023pix2video} enable editing visual concepts in an image by inverting the diffusion process and modifying cross-attentions. However, those methods require separate inference passes for each new concept and can support only a limited set of simultaneous edits. They require engineering a prompt suitable for an individual image rather than learning a simple generalizable control, and if not carefully prompted, they can introduce entanglement between concepts, such as altering race when modifying age (see Appendix). In contrast, Concept Sliders provide lightweight plug-and-play adaptors applied to pre-trained models that enable precise, continuous control over desired concepts in a single inference pass, with efficient composition (Figure~\ref{fig:2d_food}) and minimal entanglement (Figure~\ref{fig:ablations}). \par

Each Concept Slider is a low-rank modification of the diffusion model. We find that the low-rank constraint is a vital aspect of precision control over concepts: while finetuning without low-rank regularization reduces precision and generative image quality, low-rank training identifies the minimal concept subspace and results in controlled, high-quality, disentangled editing (Figure~\ref{fig:ablations}). Post-hoc image editing methods that act on single images rather than model parameters cannot benefit from this low-rank framework.\par

Concept Sliders also allow editing of visual concepts that cannot be captured by textual descriptions; this distinguishes it from prior concept editing methods that rely on text~\cite{gandikota2023erasing,gandikota2023unified}.  While image-based model customization methods~\cite{kumari2022customdiffusion, ruiz2022dreambooth, gal2022image} can add new tokens for new image-based concepts, those are difficult to use for image editing. In contrast, Concept Sliders allow an artist to provide a handful of paired images to define a desired concept, and then a Concept Slider will then generalize the visual concept and apply it to other images, even in cases where it would be infeasible to describe the transformation in words.\par 

Other generative image models, such as GANs, have previously exhibited latent spaces that provide highly disentangled control over generated outputs.  In particular, it has been observed that StyleGAN~\cite{karras2019style} stylespace neurons offer detailed control over many meaningful aspects of images that would be difficult to describe in words~\cite{wu2021stylespace}. To further demonstrate the capabilities of our approach, we show that it is possible to create Concept Sliders that transfer latent directions from StyleGAN's style space trained on FFHQ face images~\cite{karras2019style} into diffusion models. Notably, despite originating from a face dataset, our method successfully adapts these latents to enable nuanced style control over diverse image generation. This showcases how diffusion models can capture the complex visual concepts represented in GAN latents, even those that may not correspond to any textual description. \par

We demonstrate that the expressiveness of Concept Sliders is powerful enough to address two particularly practical applications---enhancing realism and fixing hand distortions. While generative models have made significant progress in realistic image synthesis, the latest generation of diffusion models such as Stable Diffusion XL~\cite{podell2023sdxl} are still prone to synthesizing distorted hands with anatomically implausible extra or missing fingers~\cite{handsreditt}, as well as warped faces, floating objects, and distorted perspectives. Through a perceptual user study, we validate that a Concept Slider for ``realistic image'' as well as another for ``fixed hands'' both create a statistically significant improvement in perceived realism without altering image content.

Concept Sliders are modular and composable. We find that over 50 unique sliders can be composed without degrading output quality. This versatility gives artists a new universe of nuanced image control that allows them to blend countless textual, visual, and GAN-defined Concept Sliders. Because our method bypasses standard prompt token limits, it empowers more complex editing than achievable through text alone. \par

\label{sec:intro}

 \section{Related Works}
\paragraph{Image Editing}
Recent methods propose different approaches for single image editing in text-to-image diffusion models. They mainly focus on manipulation of cross-attentions of a source image and a target prompt \cite{hertz2022prompt, kawar2022imagic, parmar2023zero}, or use a conditional input to guide the image structure \cite{meng2021sdedit}. 
Unlike those methods that are applied to a single image, our model creates a semantic change defined by a small set of text pairs or image pairs, applied to the entire model. Analyzing diffusion models through Riemannian geometry, Park et al. \cite{park2023understanding} discovered local latent bases that enable semantic editing by traversing the latent space. Their analysis also revealed the evolving geometric structure over timesteps across prompts, requiring per-image latent basis optimization. In contrast, we identify generalizable parameter directions, without needing custom optimization for each image. Instruct-pix2pix~\cite{brooks2022instructpix2pix} finetunes a diffusion model to condition image generation on both an input image and text prompt. This enables a wide range of text-guided editing, but lacks fine-grained control over edit strength or visual concepts not easily described textually.

\paragraph{Guidance Based Methods}
Ho et al.~\cite{ho2022classifier} introduce classifier free guidance that showed improvement in image quality and text-image alignment when the data distribution is driven towards the prompt and away from unconditional output. Liu et al.~\cite{liu2022compositional} present an inference-time guidance formulation to enhance concept composition and negation in diffusion models. By adding guidance terms during inference, their method improves on the limited inherent compositionality of diffusion models. SLD~\cite{schramowski2022safe} proposes using guidance to moderate unsafe concepts in diffusion models. They propose a safe prompt which is used to guide the output away from unsafe content during inference. 

\paragraph{Model Editing}
Our method can be seen as a model editing approach, where by applying a low-rank adaptor, we single out a semantic attribute and allow for continuous control with respect to the attribute. To personalize the models for adding new concepts, customization methods based on finetuning exist~\cite{ruiz2022dreambooth, kumari2022customdiffusion, gal2022image}. Custom Diffusion~\cite{kumari2022customdiffusion} proposes a way to incorporate new visual concepts into pretrained diffusion models by finetuning only the cross-attention layers. On the other hand, Textual Inversion~\cite{gal2022image} introduces new textual concepts by optimizing an embedding vector to activate desired model capabilities.
Previous works~\cite{gandikota2023erasing, kumari2023conceptablation, kim2023towards,heng2023selective,zhang2023forget} proposed gradient based fine-tuning-based methods for the permanent erasure of a concept in a model. Ryu et al.~\cite{Ryu_lora} proposed adapting LoRA~\cite{hu2021lora} for diffusion model customization. Recent works ~\cite{Zhou_leco} developed low rank implementations of erasing concepts~\cite{gandikota2023erasing} allowing the ability to adjust the strength of erasure in an image. \cite{Inui_2023} implemented image based control of concepts by merging two overfitted LoRAs to capture an edit direction. Similarly, ~\cite{gandikota2023unified, orgad2023editing} proposed closed-form formulation solutions for debiasing, redacting or moderating concepts within the model's cross-attention weights. Our method does not modify the underlying text-to-image diffusion model and can be applied as a plug-and-play module easily stacked across different attributes. %

\paragraph{Semantic Direction in Generative models}
In Generative Adversarial Networks (GANs), manipulation of semantic attributes has been widely studied. Latent space trajectories have been found in a self-supervised manner~\cite{jahanian2019steerability}. PCA has been used to identify semantic directions in the latent or feature spaces~\cite{harkonen2020ganspace}. Latent subspaces corresponding to detailed face attributes have been analyzed~\cite{Shen_2020_CVPR}. For diffusion models, semantic latent spaces have been suggested to exist in the middle layers of the U-Net architecture~\cite{kwon2022diffusion, park2023unsupervised}. It has been shown that principal directions in diffusion model latent spaces (h-spaces) capture global semantics~\cite{haas2023discovering}. Our method directly trains low-rank subspaces corresponding to semantic attributes. By optimizing for specific global directions using text or image pairs as supervision, we obtain precise and localized editing directions. Recent works have \cite{zou2023representation} introduced the low-rank representation adapter, which employs a contrastive loss to fine-tune LoRA to achieve fine-grained control of concepts in language models.

 \section{Background}
\subsection{Diffusion Models}
Diffusion models are a subclass of generative models that operationalize the concept of reversing a diffusion process to synthesize data. Initially, the forward diffusion process gradually adds noise to the data, transitioning it from an organized state $x_0$ to a complete Gaussian noise $x_T$. At any timestep $t$, the noised image is modelled as:
\begin{equation}
\label{eq:forward}
x_t \gets \sqrt{1-\beta_t}x_0 + \sqrt{\beta_t}\epsilon
\end{equation}
Where $\epsilon$ is a randomly sampled gaussian noise with zero mean and unit variance. Diffusion models aim to reverse this diffusion process by sampling a random Gaussian noise $X_T$ and gradually denoising the image to generate an image $x_0$. In practice \cite{ho2020denoising, luo2022understanding}, the objective of diffusion model is simplified to predicting the true noise $\epsilon$ from Eq.~\ref{eq:forward} when $x_t$ is fed as input with additional inputs like the timestep $t$ and conditioning $c$.
\begin{equation}
    \label{eq:normal-diffusion}
    \nabla_\theta||\epsilon - \epsilon_\theta(x_t, c, t)||^2
\end{equation}
Where $\epsilon_\theta(x_t, c, t)$ is the noise predicted by the diffusion model conditioned on $c$ at timestep $t$. In this work, we work with Stable Diffusion~\cite{rombach2022high} and Stable Diffusion XL~\cite{podell2023sdxl}, which are latent diffusion models that improve efficiency by operating in a lower dimensional latent space $z$ of a pre-trained variational autoencoder. They convert the images to a latent space and run the diffusion training as discussed above. Finally, they decode the latent $z_0$ through the VAE decoder to get the final image $x_0$

\subsection{Low-Rank Adaptors}
The Low-Rank Adaptation (LoRA)~\cite{hu2021lora} method enables efficient adaptation of large pre-trained language models to downstream tasks by decomposing the weight update $\Delta W$ during fine-tuning. Given a pre-trained model layer with weights $W_0 \in \mathbb{R}^{d \times k}$, where $d$ is the input dimension and $k$ the output dimension, LoRA decomposes $\Delta W$ as
\begin{equation}
\Delta W = BA
\end{equation}
where $B \in \mathbb{R}^{d \times r}$ and $A \in \mathbb{R}^{r \times k}$ with $r \ll \min(d,k)$ being a small rank that constrains the update to a low dimensional subspace. By freezing $W_0$ and only optimizing the smaller matrices $A$ and $B$, LoRA achieves massive reductions in trainable parameters.
During inference, $\Delta W$ can be merged into $W_0$ with no overhead by a LoRA scaling factor $\alpha$:
\begin{equation}
\label{eq:lora}
W = W_0 + \alpha \Delta W
\end{equation}

\begin{figure}
    \centering
    \includegraphics[width=\linewidth]{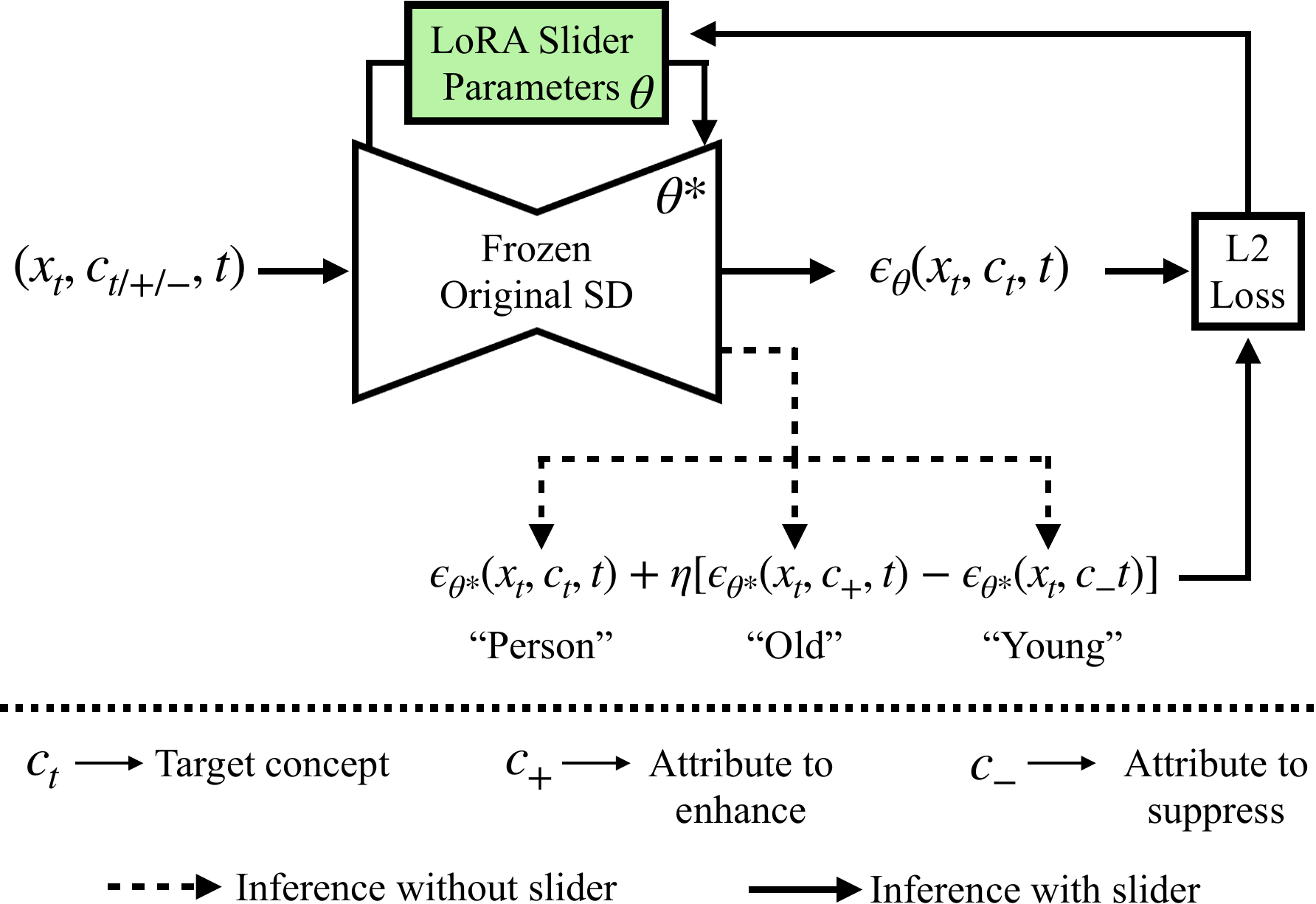}
    \caption{Concept Sliders are created by fine-tuning LoRA adaptors using a guided score that enhances attribute $c_+$ while suppressing attribute $c_-$ from the target concept $c_t$. The slider model generates samples $x_t$ by partially denoising Gaussian noise over time steps 1 to $t$, conditioned on the target concept $c_t$. }
    \label{fig:method}
\end{figure}
\section{Method}
Concept Sliders are a method for fine-tuning LoRA adaptors on a diffusion model to enable concept-targeted image control as shown in Figure~\ref{fig:method}. Our method learns low-rank parameter directions that increase or decrease the expression of specific attributes when conditioned on a target concept. Given a target concept $c_t$ and model $\theta$, our goal is to obtain $\theta^*$ that modifies the likelihood of attributes $c_+$ and $c_-$ in image $X$ when conditioned on $c_t$ - increase likelihood of attribute $c_+$ and decrease likelihood of attribute $c_-$.
\begin{align}
    P_{\theta^*}(X | c_t) \gets P_{\theta}(X | c_t) \left(\frac{ P_{\theta}(c_+ | X)}{ P_{\theta}(c_- | X)}\right)^{\eta}
\end{align}
Where $P_{\theta}(X | c_t)$ represents the distribution generated by the original model when conditioned on $c_t$. Expanding $P(c_+|X)=\frac{P(X|c_+)P(c_+)}{P(X)}$, the gradient of the log probability $\nabla \log P_{\theta^*}(X|c_t)$ would be proportional to:
\begin{align}
\label{eq:gradlogP}
    \nabla \log P_{\theta}(X | c_t) + \eta \left(\nabla \log P_{\theta}(X|c_+) - \nabla \log P_{\theta}(X|c_-)\right)
\end{align}

Based on Tweedie's formula \cite{efron2011tweedie} and the reparametrization trick of \cite{ho2020denoising},  we can introduce a time-varying noising process and express each score (gradient of log probability) as a denoising prediction $\epsilon(X,c_t,t)$. Thus Eq.~\ref{eq:gradlogP} becomes:
\begin{align}
\label{eq:objective}
\begin{split}
    \epsilon_{\theta^*}(X, c_t, t)  \gets \; &  \epsilon_{\theta}(X, c_t, t) \; + \\
    & \eta\left(\epsilon_{\theta}(X, c_+, t) - \epsilon_{\theta}(X, c_-, t) \right)
\end{split}
\end{align}
The proposed score function in Eq. \ref{eq:objective} shifts the distribution of the target concept $c_t$ to exhibit more attributes of $c_+$ and fewer attributes of $c_-$. In practice, we notice that a single prompt pair can sometimes identify a direction that is entangled with other undesired attributes. We therefore incorporate a set of preservation concepts $p\in \mathcal{P}$ (for example, race names while editing age) to constrain the optimization. Instead of simply increasing $P_{\theta}(c_+|X)$, we aim to increase, for every $p$, $P_{\theta}((c_+, p)|X)$, and reduce  $P_{\theta}((c_-, p)|X)$. This leads to the disentanglement objective:
\begin{align}
\label{eq:objective_disentangle}
\begin{split}
    \epsilon_{\theta^*}(X, c_t, t)  & \gets \;  \epsilon_{\theta}(X, c_t, t) \; + \\ 
    & \eta \sum_{p\in \mathcal{P}} \left(\epsilon_{\theta}(X, (c_+, p), t) - \epsilon_{\theta}(X, (c_-,p), t)\right)
\end{split}
\end{align}
The disentanglement objective in Equation~\ref{eq:objective_disentangle} finetunes the Concept Slider modules while keeping pre-trained weights fixed. Crucially, the LoRA formulation in Equation~\ref{eq:lora} introduces a scaling factor $\alpha$ that can be modified at inference time. This scaling parameter $\alpha$ allows adjusting the strength of the edit, as shown in Figure~\ref{fig:intro}. Increasing $\alpha$ makes the edit stronger without retraining the model. Previous model editing method \cite{gandikota2023erasing}, suggests a stronger edit by retraining with increased guidance $\eta$ in Eq.~\ref{eq:objective_disentangle}. However, simply scaling $\alpha$ at inference time produces the same effect of strengthening the edit, without costly retraining.%

\subsection{Learning Visual Concepts from Image Pairs}
We propose sliders to control nuanced visual concepts that are harder to specify using text prompts. We leverage small paired before/after image datasets to train sliders for these concepts. The sliders learn to capture the visual concept through the contrast between image pairs ($x^A$, $x^B$). \par
Our training process optimizes the LORA applied in both the negative and positive directions.  We shall write $\epsilon_{\theta_+}$ for the application of positive LoRA and $\epsilon_{\theta_-}$ for the negative case. 
Then we minimize the following loss:
\begin{equation}
   ||\epsilon_{\theta_-}(x^A_t, \text{` '}, t) - \epsilon||^2 + ||\epsilon_{\theta_+}(x^B_t, \text{` '}, t) - \epsilon||^2 %
\end{equation}
This has the effect of causing the LORA to align to a direction that causes the visual effect of A in the negative direction and B in the positive direction.  Defining directions visually in this way not only allows an artist to define a Concept Slider through custom artwork; it is also the same method we use to transfer latents from other generative models such as StyleGAN.

 \begin{figure}
    \centering
    \includegraphics[width=\linewidth]{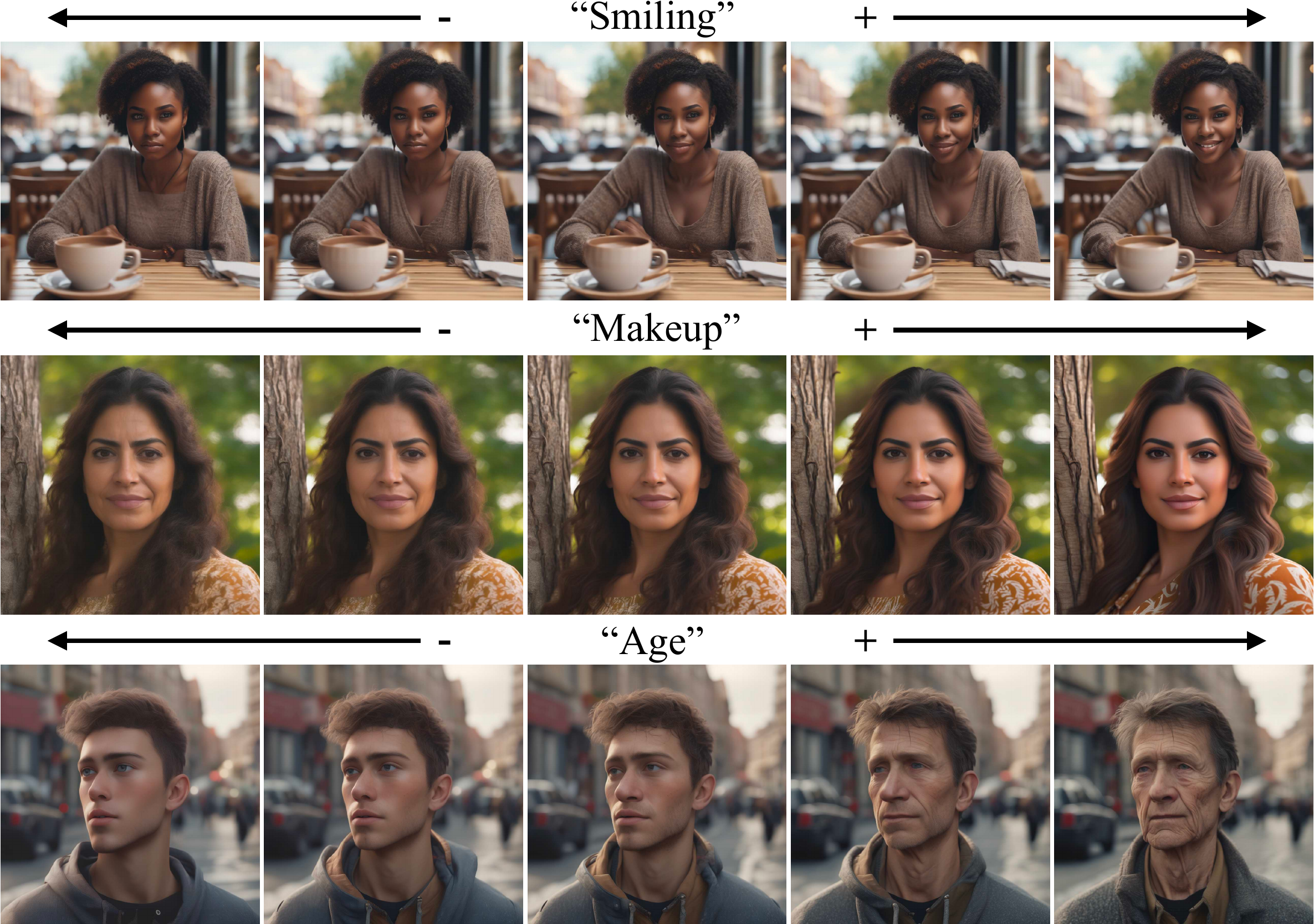}
    \caption{Our text-based sliders allow precise editing of desired attributes during image generation while maintaining the overall structure. Traversing the sliders towards the negative direction produces an opposing effect on the attributes.}
    \label{fig:textslider}
\end{figure}
\section{Experiments}
We evaluate our approach primarily on Stable Diffusion XL \cite{podell2023sdxl}, a high-resolution 1024-pixel  model, and we conduct additional experiments on SD v1.4 \cite{rombach2022high}. All models are trained for 500 epochs. We demonstrate generalization by testing sliders on diverse prompts - for example, we evaluate our "person" slider on prompts like "doctor", "man", "woman", and "barista". For inference, we follow the SDEdit technique of Meng et al.~\cite{meng2021sdedit}: to maintain structure and semantics, we use the original pre-trained model for the first $t$ steps, setting the LoRA adaptor multipliers to 0 and retaining the pre-trained model priors. We then turn on the LoRA adaptor for the remaining steps.

\subsection{Textual Concept Sliders}
\label{sec:textsliders}

We validate the efficacy of our slider method on a diverse set of 30 text-based concepts, with full examples in the Appendix. Table~\ref{tab:textslider} compares our method against two baselines: an approach we propose inspired by SDEdit~\cite{meng2021sdedit} and Liu et al.\cite{liu2022compositional} that uses a pretrained model with the standard prompt for $t$ timesteps, then starts composing by adding prompts to steer the image, and prompt2prompt\cite{hertz2022prompt}, which leverages cross-attention for image editing after generating reference images. While the former baseline is novel, all three enable finer control but differ in how edits are applied. Our method directly generates 2500 edited images per concept, like "image of a person", by setting the scale parameter at inference. In contrast, the baselines require additional inference passes for each new concept (e.g "old person"), adding computational overhead. Our method consistently achieves higher CLIP scores and lower LPIPS versus the original, indicating greater coherence while enabling precise control. The baselines are also more prone to entanglement between concepts. We provide further analysis and details about the baselines in the Appendix.\par
Figure~\ref{fig:textslider} shows typical qualitative examples,
which maintains good image structure while enabling fine
grained editing of the specified concept.

\begin{table}
\centering
\resizebox{\columnwidth}{!}{%
\begin{tabular}{lcccccc}
& \multicolumn{2}{c}{\textbf{Prompt2Prompt}}  & \multicolumn{2}{c}{\textbf{Our Method}} & \multicolumn{2}{c}{\textbf{Composition}} \tabularnewline
& $\Delta$ CLIP & LPIPS & $\Delta$ CLIP & LPIPS & $\Delta$ CLIP & LPIPS \tabularnewline
\hline
Age   & 1.10 & 0.15 & \textbf{3.93} & \textbf{0.06} & 3.14 & 0.13  \tabularnewline %
Hair  & 3.45 & 0.15 & \textbf{5.59} & \textbf{0.10} & 5.14 & 0.15 \tabularnewline %
Sky   & 0.43 & 0.15 & \textbf{1.56}  & \textbf{0.13} & 1.55 & 0.14 \tabularnewline
Rusty   & \textbf{7.67} & 0.25 & 7.60 & \textbf{0.09} & 6.67 & 0.18
\end{tabular}
}
\caption{Compared to Prompt2Prompt~\cite{hertz2022prompt}, our method achieves comparable efficacy in terms of $\Delta$ CLIP score while inducing finer edits as measured by LPIPS distance to the original image. The $\Delta$ CLIP metric measures the change in CLIP score between the original and edited images when evaluated on the text prompt describing the desired edit. Results are shown for a single positive scale of the trained slider.}

\label{tab:textslider}
\end{table}

\subsection{Visual Concept Sliders}
\begin{figure}
    \centering
    \includegraphics[width=\linewidth]{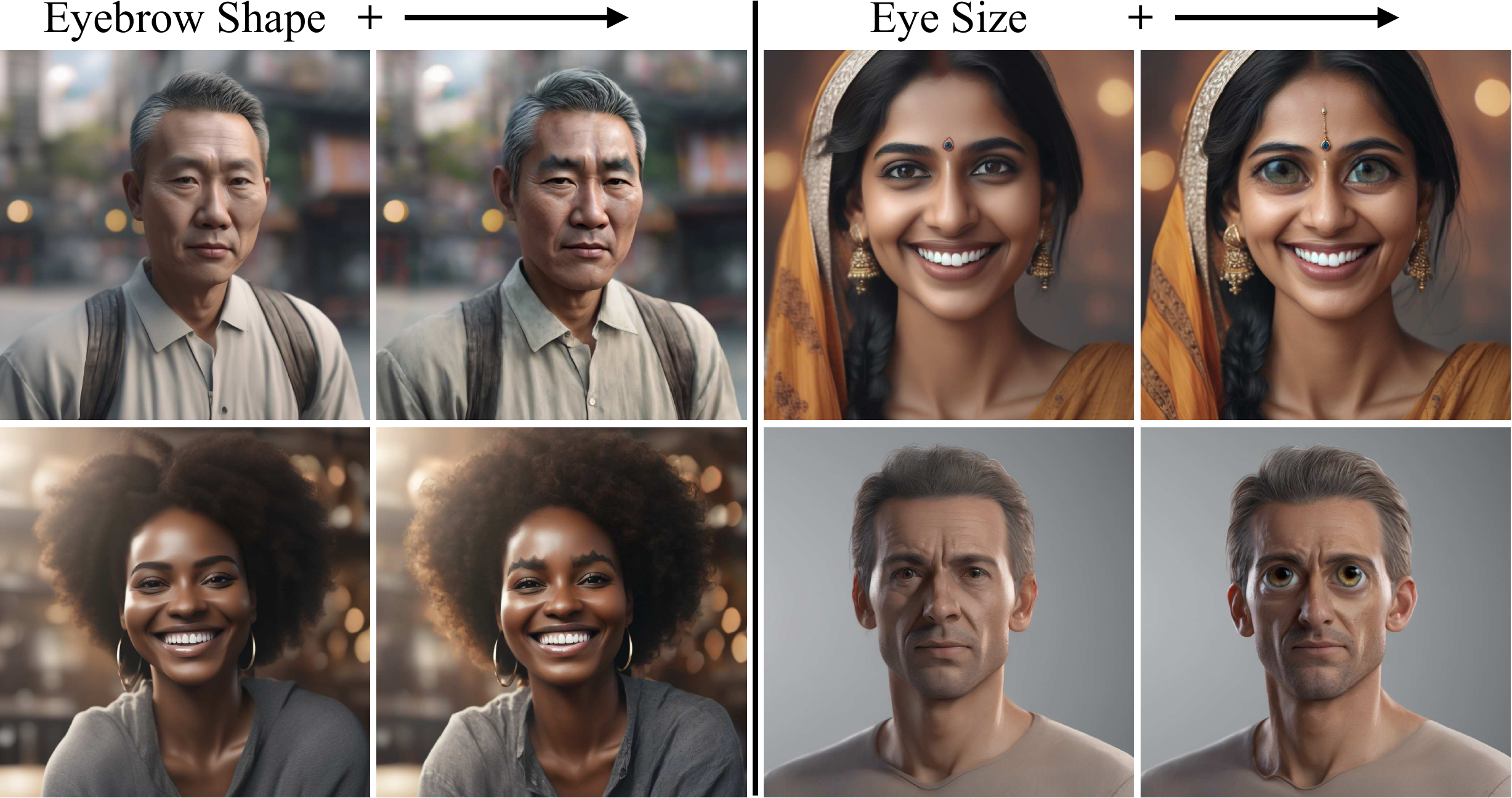}
    \caption{Controlling fine-grained attributes like eyebrow shape and eye size using image pair-driven concept sliders with optional text guidance. The eye size slider scales from small to large eyes using the Ostris dataset \cite{ostris}.}
    \label{fig:imageslider}
\end{figure}

Some visual concepts like precise eyebrow shapes or eye sizes are challenging to control through text prompts alone. To enable sliders for these granular attributes, we leverage paired image datasets combined with optional text guidance. As shown in Figure \ref{fig:imageslider}, we create sliders for "eyebrow shape" and "eye size" using image pairs capturing the desired transformations. We can further refine the eyebrow slider by providing the text "eyebrows" so the direction focuses on that facial region. Using image pairs with different scales, like the eye sizes from Ostris \cite{ostris}, we can create sliders with stepwise control over the target attribute. \par
We quantitatively evaluate the eye size slider by detecting faces using FaceNet~\cite{schroff2015facenet}, cropping the area, and employing a face parser~\cite{faceparser}to measure eye region across the slider range. Traversing the slider smoothly increases the average eye area 2.75x, enabling precise control as shown in Table \ref{tab:imageslider}. Compared to customization techniques like textual inversion~\cite{gal2022image} that learns a new token and custom diffusion~\cite{kumari2022customdiffusion} that fine-tunes cross attentions, our slider provides more targeted editing without unwanted changes. When model editing methods~\cite{kumari2022customdiffusion,gal2022image} are used to incorporate new visual concepts, they memorize the training subjects rather than generalizing the contrast between pairs. We provide more details in the Appendix.
\begin{table}[]
    \centering
    \begin{tabular}{rcccc}
         & \hspace{-1em} \textbf{Training} & \textbf{Custom }& \textbf{Textual} & \textbf{Our}  \\
        & \hspace{-1em}\textbf{Data} & \textbf{Diffusion }& \textbf{Inversion} & \textbf{Method}  \\
        \hline
        $\mathbf{\Delta_{eye}}$ & \hspace{-1em}1.84 & 0.97 & 0.81 & \textbf{1.75} \\
        \textbf{LPIPS} &\hspace{-1em} 0.03 & 0.23 & 0.21 & \textbf{0.06} 
    \end{tabular}
    \caption{Our results demonstrate the effectiveness of our sliders for intuitive image editing based on visual concepts. The metric $\Delta_{eye}$ represents the ratio of change in eye size compared to the original image. Our method achieves targeted editing of eye size while maintaining similarity to the original image distribution, as measured by the LPIPS.}
    \label{tab:imageslider}
\end{table}
\subsection{Sliders transferred from StyleGAN}
\begin{figure}
    \centering
    \includegraphics[width=\linewidth]{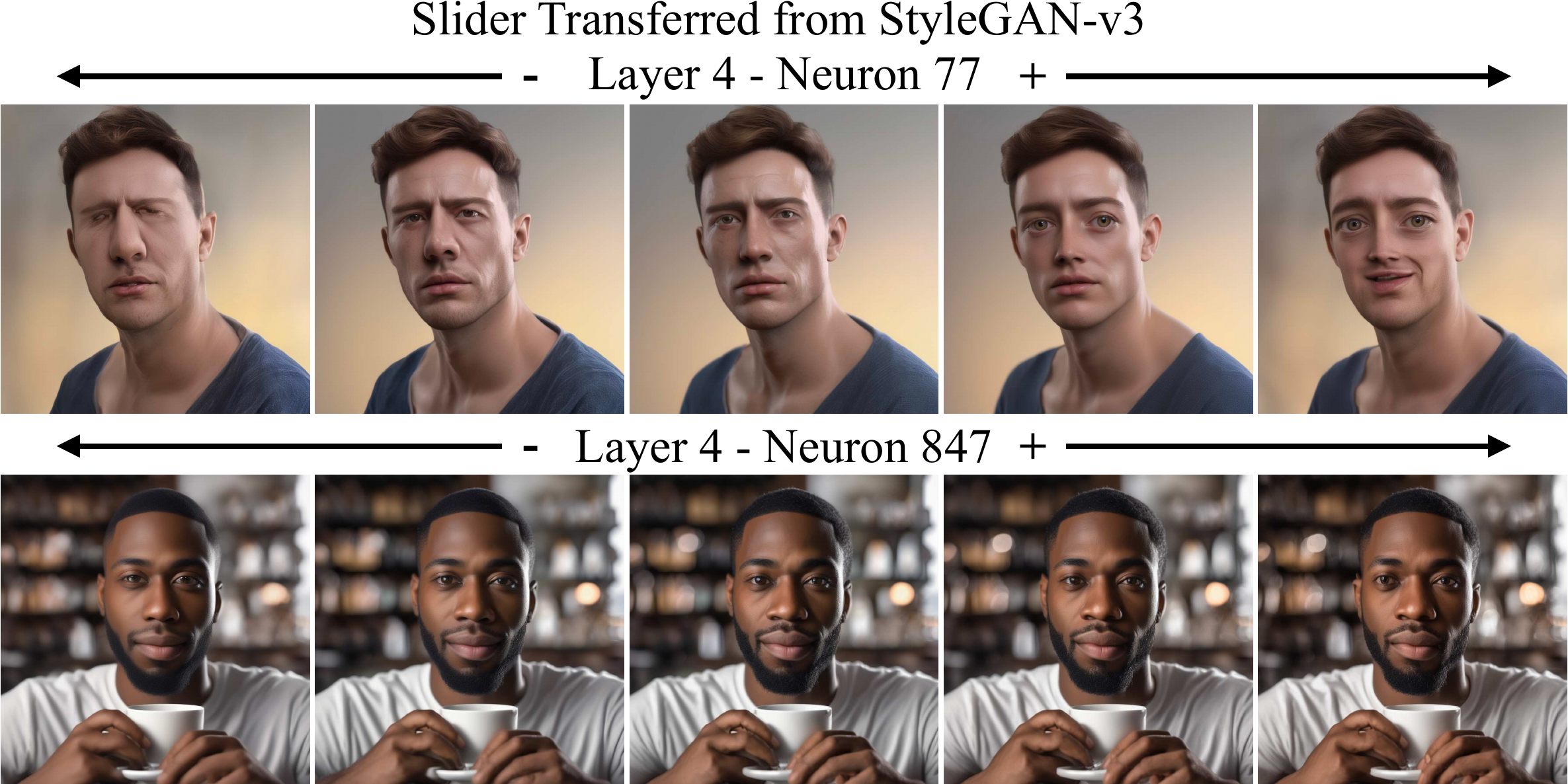}
    \caption{We demonstrate transferring StyleGAN style space latents to the diffusion latent space. We identify three neurons that edit facial structure: neuron 77 controls cheekbone structure, neuron 646 selectively adjusts the left side face width, and neuron 847 edits inter-ocular distance. We transfer these StyleGAN latents to the diffusion model to enable structured facial editing.}
    \label{fig:ganslider}
\end{figure}

Figure~\ref{fig:ganslider} demonstrates sliders transferred from the StyleGAN-v3~\cite{karras2021alias} style space that is trained on FFHQ~\cite{karras2019style} dataset. We use the method of~\cite{wu2021stylespace} to explore the StyleGAN-v3 style space and identify neurons that control hard-to-describe facial features. By scaling these neurons, we collect images to train image-based sliders. We find that Stable Diffusion's latent space can effectively learn these StyleGAN style neurons, enabling structured facial editing. This enables users to control nuanced concepts that are indescribable by words and styleGAN makes it easy to get generate the paired dataset.
\subsection{Composing Sliders}
\begin{figure}
    \centering
    \includegraphics[width=\linewidth]{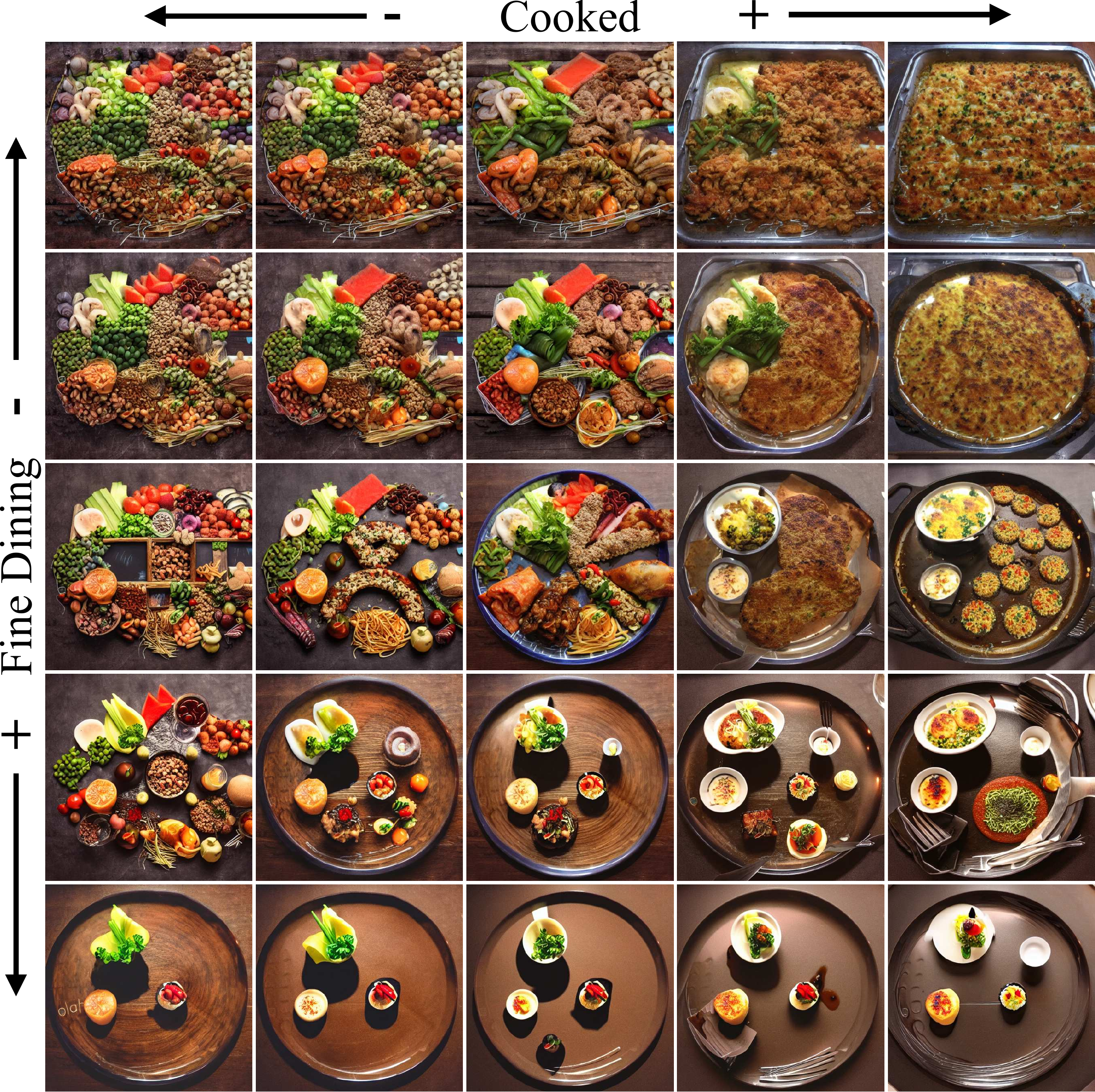}
    \caption{Composing two text-based sliders results in a complex control over food images. We show the effect of applying both the "cooked" slider and "fine-dining" slider to a generated image. These sliders can be used in both positive and negative directions. }
    \label{fig:2d_food}
\end{figure}

\begin{figure}
    \centering
    \includegraphics[width=\linewidth]{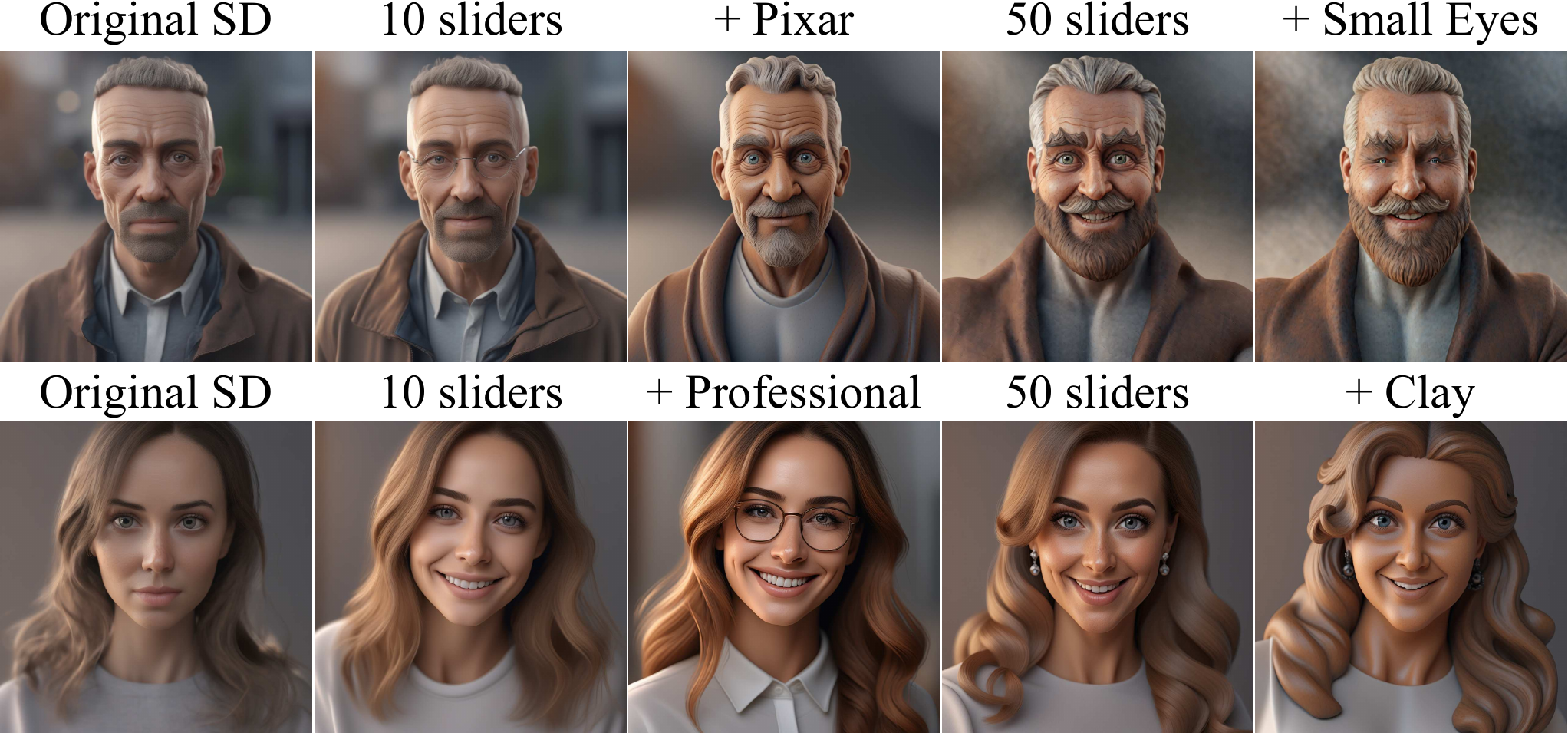}
    \caption{We show composition capabilities of concept sliders. We progressively compose multiple sliders in each row from left to right, enabling nuanced traversal of high-dimensional concept spaces. We demonstrate composing sliders trained from text prompts, image datasets, and transferred from GANs.}
    \label{fig:composable-limit}
\end{figure}

A key advantage of our low-rank slider directions is composability - users can combine multiple sliders for nuanced control rather than being limited to one concept at a time. For example, in Figure \ref{fig:2d_food} we show blending "cooked" and "fine dining" food sliders to traverse this 2D concept space. Since our sliders are lightweight LoRA adaptors, they are easy to share and overlay on diffusion models. By downloading interesting slider sets, users can adjust multiple knobs simultaneously to steer complex generations. In Figure~\ref{fig:composable-limit} we qualitatively show the effects of composing multiple sliders progressively up to 50 sliders at a time. We use far greater than 77 tokens (the current context limit of SDXL~\cite{podell2023sdxl}) to create these 50 sliders. This showcases the power of our method that allows control beyond what is possible through prompt-based methods alone. We further validate multi-slider composition in the appendix. 

\begin{figure*}
    \centering
    \includegraphics[width=\linewidth]{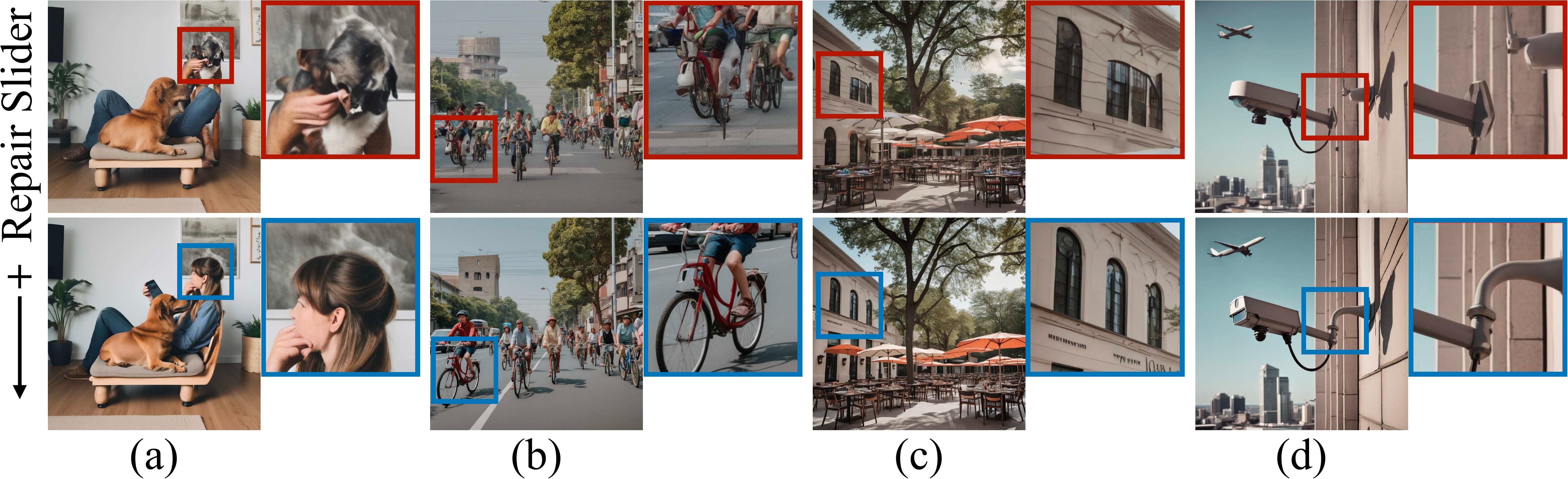}
    \caption{The repair slider enables the model to generate images that are more realistic and undistorted. The parameters under the control of this slider help the model correct some of the flaws in their generated outputs like distorted humans and pets in (a, b), unnatural objects in (b, c, d), and blurry natural images in (b,c)}
    \label{fig:realistic}
\end{figure*}

\section{Concept Sliders to Improve Image Quality}

One of the most interesting aspects of a large-scale generative model such as Stable Diffusion XL is that, although their image output can often suffer from distortions such as warped or blurry objects, the parameters of the model contains a latent capability to generate higher-quality output with fewer distortions than produced by default.  Concept Sliders can unlock these abilities by identifying low-rank parameter directions that repair common distortions.

\begin{figure}
    \centering
    \includegraphics[width=\linewidth]{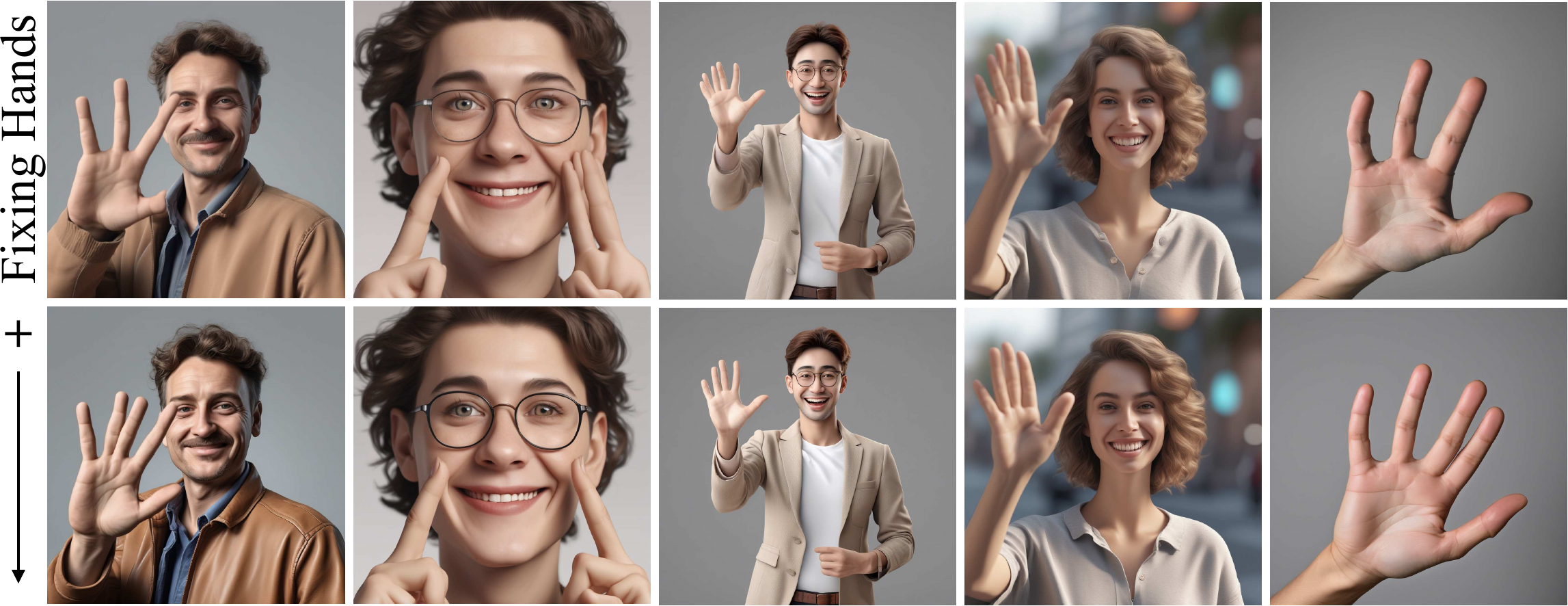}
    \caption{We demonstrate a slider for fixing hands in stable diffusion. We find a direction to steer hands to be more realistic and away from "poorly drawn hands".}
    \label{fig:fixinghands}
\end{figure}

\paragraph{Fixing Hands}
Generating realistic-looking hands is a persistent challenge for diffusion models: for example, hands are typically generated with missing, extra, or misplaced fingers. Yet the tendency to distort hands can be directly controlled by a Concept Slider: Figure~\ref{fig:fixinghands} shows the effect of a "fix hands" Concept Slider that lets users smoothly adjust images to have more realistic, properly proportioned hands. This parameter direction is found using a complex prompt pair boosting ``realistic hands, five fingers, 8k hyper-realistic hands'' and suppressing ``poorly drawn hands, distorted hands, misplaced fingers''. This slider allows hand quality to be improved with a simple tweak rather manual prompt engineering. \par

To measure the ``fix hands" slider, we conduct a user study on Amazon Mechanical Turk. We present 300 random images with hands to raters---half generated by Stable Diffusion XL and half by XL with our slider applied (same seeds and prompts). Raters are asked to assess if the hands appear distorted or not. Across 150 SDXL images, raters find 62\% have distorted hands, confirming it as a prevalent problem. In contrast, only 22\% of the 150 slider images are rated as having distorted hands. 

\begin{figure}
    \centering
    \includegraphics[width=\linewidth]{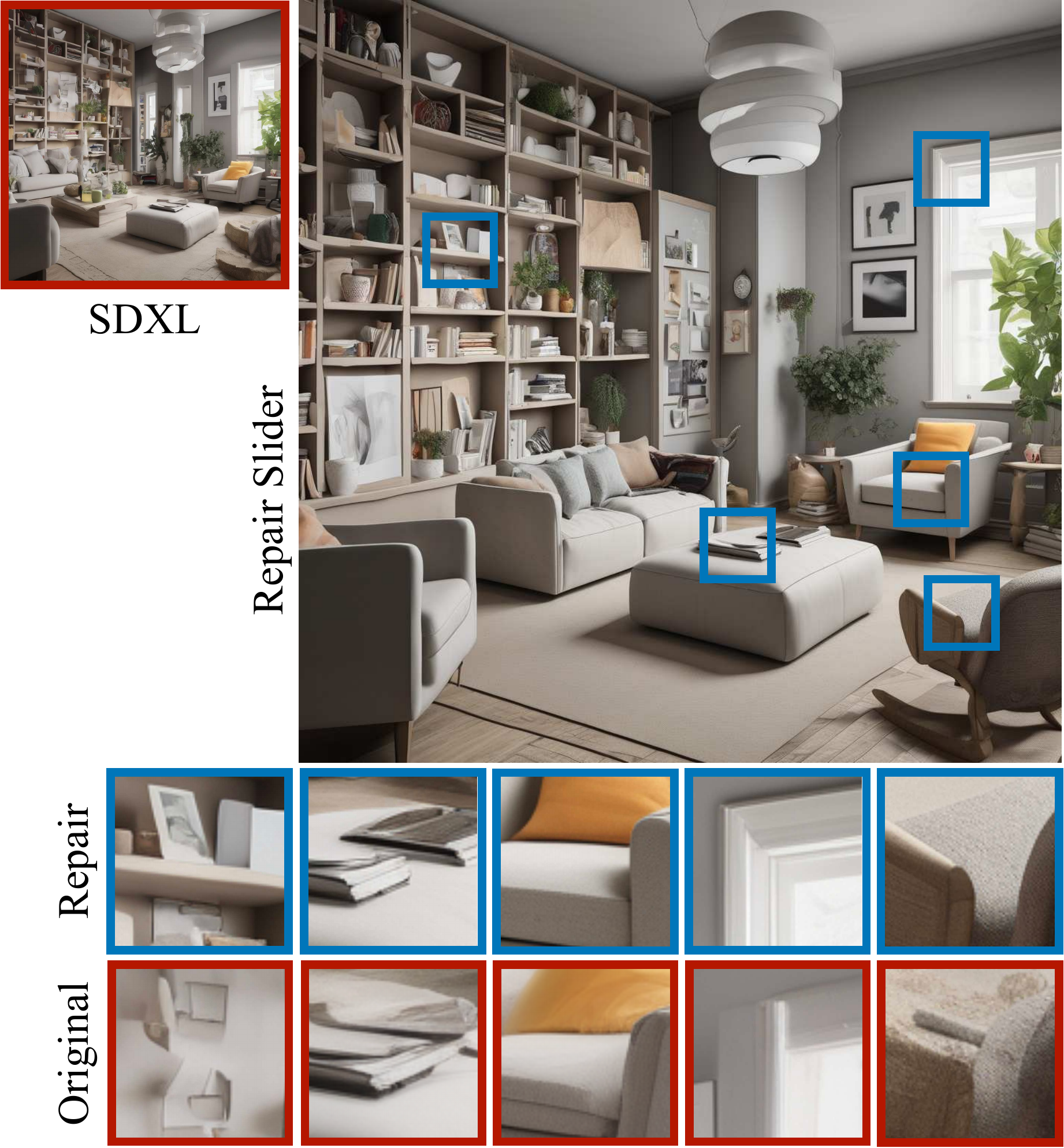}
    \caption{We demonstrate the effect of our ``repair'' slider on fine details: it improves the rendering of densely arranged objects, it straightens architectural lines, and it avoids blurring and distortions at the edges of complex shapes.}
    \label{fig:realistic_large}
\end{figure}

\paragraph{Repair Slider}
In addition to controlling specific concepts like hands, we also demonstrate the use of Concept Sliders to guide generations towards overall greater realism. We identify single low-rank parameter direction that shifts images away from common quality issues like distorted subjects, unnatural object placement, and inconsistent shapes. As shown in Figures~\ref{fig:realistic} and~\ref{fig:realistic_large}, traversing this ``repair" slider noticeably fixes many errors and imperfections.

Through a perceptual study, we evaluate the realism of 250 pairs of slider-adjusted and original SD images. A majority of participants rate the slider images as more realistic in 80.39\% of pairs, indicating our method enhances realism. However, FID scores do not align with this human assessment, echoing prior work on perceptual judgment gaps \cite{kynkaanniemi2022role}. Instead, distorting images along the opposite slider direction improves FID, though users still prefer the realism-enhancing direction. We provide more details about the user studies in the appendix.

\section{Ablations}
\begin{figure}[ht]
    \centering
    \includegraphics[width=\linewidth]{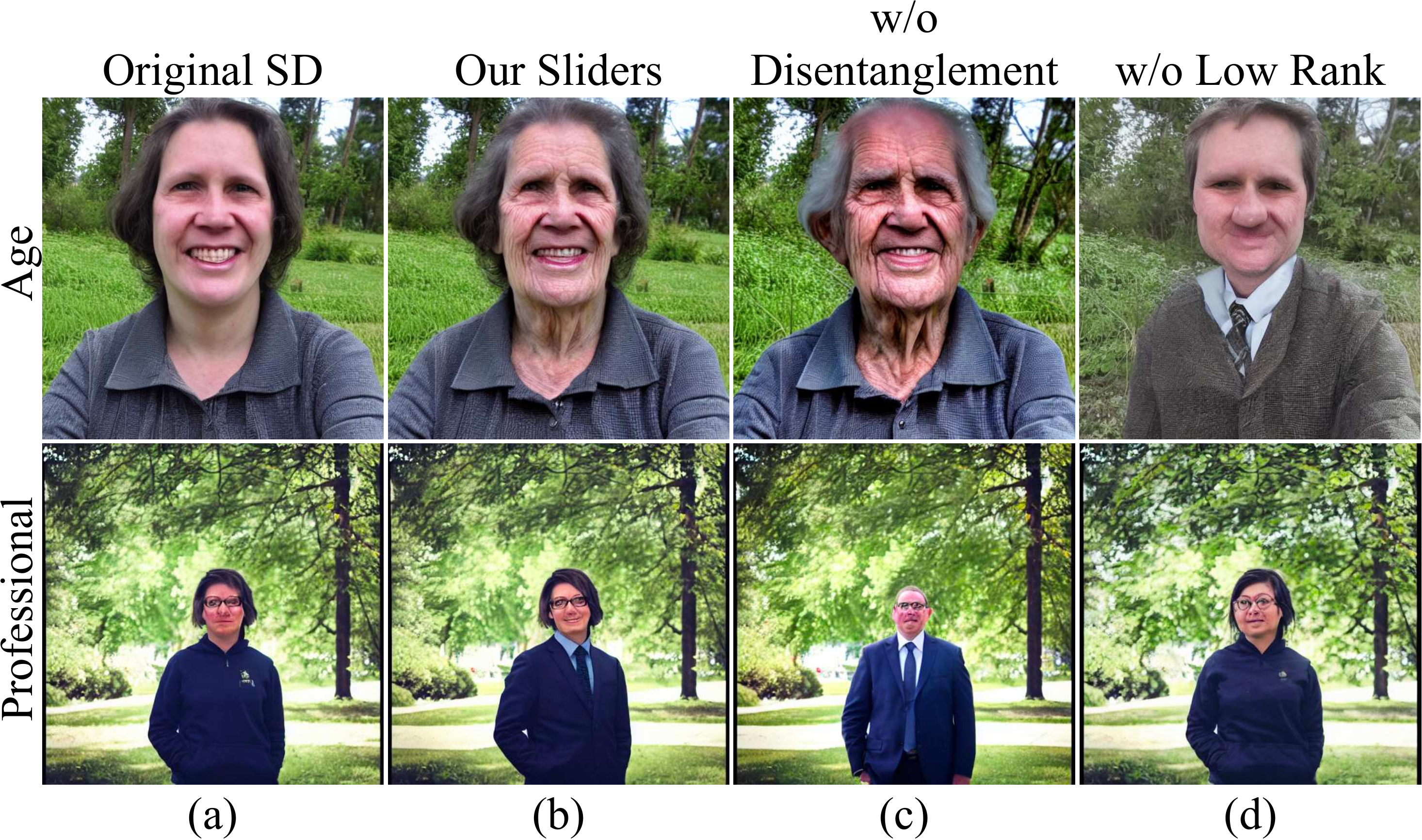}
    \caption{The disentanglement objective (Eq.~\ref{eq:objective_disentangle}) helps avoid undesired attribute changes like change in race or gender when editing age. The low-rank constraint enables a precise edit.}
    \label{fig:ablations}
\end{figure}

We analyze the two key components of our method to verify that they are both necessary: (1) the disentanglement formulation and (2) low-rank adaptation. Table~\ref{tab:ablations} shows quantitative measures on 2500 images, and Figure \ref{fig:ablations} shows qualitative differences. In both quantitative and quantitative measures, we find that the disentanglement objective from Eq.\ref{eq:objective_disentangle} success in isolating the edit from unwanted attributes (Fig.\ref{fig:ablations}.c); for example without this objective we see undesired changes in gender when asking for age as seen in Table~\ref{tab:ablations}, Interference metric which measures the percentage of samples with changed race/gender when making the edit. The low-rank constraint is also helpful: it has the effect of precisely capturing the edit direction with better generalization (Fig.\ref{fig:ablations}.d); for example, note how the background and the clothing are better preserved in Fig.\ref{fig:ablations}.b. Since LORA is parameter-efficient, it also has the advantage that it enables lightweight modularity. We also note that the SDEdit-inspired inference technique allows us to use a wider range of alpha values, increasing the editing capacity, without losing image structure. We find that SDEdit's inference technique expands the usable range of alpha before coherence declines relative to the original image. We provide more details in the Appendix.

\begin{table}[ht]
    \centering
    \begin{tabular}{rccc}
    &&  \textbf{w/o}& \textbf{w/o} \\
    &\textbf{Ours}& \hspace{-.5em}\textbf{Disentanglement}
    &\hspace{-.5em}\textbf{Low Rank}  \\
    \hline
    $\mathbf{\Delta_{CLIP}}$   & \textbf{3.93} &\hspace{-.5em} 3.39 &\hspace{-.5em}3.18 \\ 
    \textbf{LPIPS }           &\textbf{0.06}&\hspace{-.5em}0.17&\hspace{-.5em}0.23\\
    \textbf{Interference}     &\textbf{0.10}&\hspace{-.5em}0.36&\hspace{-.5em}0.19\\
    \end{tabular}
    \caption{The disentanglement formulation enables precise control over the age direction, as shown by the significant reduction in the Interference metric which measures the percentage of samples with gender/race change, compared to the original images. By using LoRA adaptors, sliders achieve finer editing in terms of both structure and edit direction, as evidenced by improvements in LPIPS and Interference. Concept strength is maintained, with similar $\Delta_{CLIP}$ scores across ablations.}
    \label{tab:ablations}
\end{table}

 \section{Limitations}
While the disentanglement formulation reduces unwanted interference between edits, we still observe some residual effects as shown in Table~\ref{tab:ablations} for our sliders. This highlights the need for more careful selection of the latent directions to preserve, preferably an automated method, in order to further reduce edit interference. Further study is required to determine the optimal set of directions that minimizes interference while retaining edit fidelity. We also observe that while the inference SDEdit technique helps preserve image structure, it can reduce edit intensity compared to the inference-time method, as shown in Table~\ref{tab:textslider}. The SDEdit approach appears to trade off edit strength for improved structural coherence. Further work is needed to determine if the edit strength can be improved while maintaining high fidelity to the original image.
 \section{Conclusion}

Concept Sliders are a simple and scalable new paradigm for interpretable control of diffusion models. By learning precise semantic directions in latent space, sliders enable intuitive and generalized control over image concepts. The approach provides a new level of flexiblilty beyond text-driven, image-specific diffusion model editing methods, because Concept Sliders allow continuous, single-pass adjustments without extra inference. Their modular design further enables overlaying many sliders simultaneously, unlocking complex multi-concept image manipulation. \par

We have demonstrated the versatility of Concept Sliders by measuring their performance on Stable Diffusion XL and Stable Diffusion 1.4.  We have found that sliders can be created from textual descriptions alone to control abstract concepts with minimal interference with unrelated concepts, outperforming previous methods. We have demonstrated and measured the efficacy of sliders for nuanced visual concepts that are difficult to describe by text, derived from small artist-created image datasets. We have shown that Concept Sliders can be used to transfer StyleGAN latents into diffusion models. Finally, we have conducted a human study that verifies the high quality of Concept Sliders that enhance and correct hand distortions. Our code and data will be made publicly available.

\section*{Acknowledgments}
We thank Jaret Burkett (aka Ostris) for the continued discussion on the image slider method and for sharing their eye size dataset. RG and DB are supported by Open Philanthropy. 
\section*{Code}
 Our methods are available as open-source code. Source code, trained sliders, and data sets for reproducing our results can be found at \href{https://sliders.baulab.info/}{sliders.baulab.info} and at \href{https://github.com/rohitgandikota/sliders}{https://github.com/rohitgandikota/sliders}.
 
{
    \small
    \bibliographystyle{ieeenat_fullname}
    \bibliography{main}
}

 \clearpage
\setcounter{page}{1}
\maketitlesupplementary

\section{Disentanglement Formulation}
\label{sec:rationale}
\begin{figure}
    \centering
    \includegraphics[width=\linewidth]{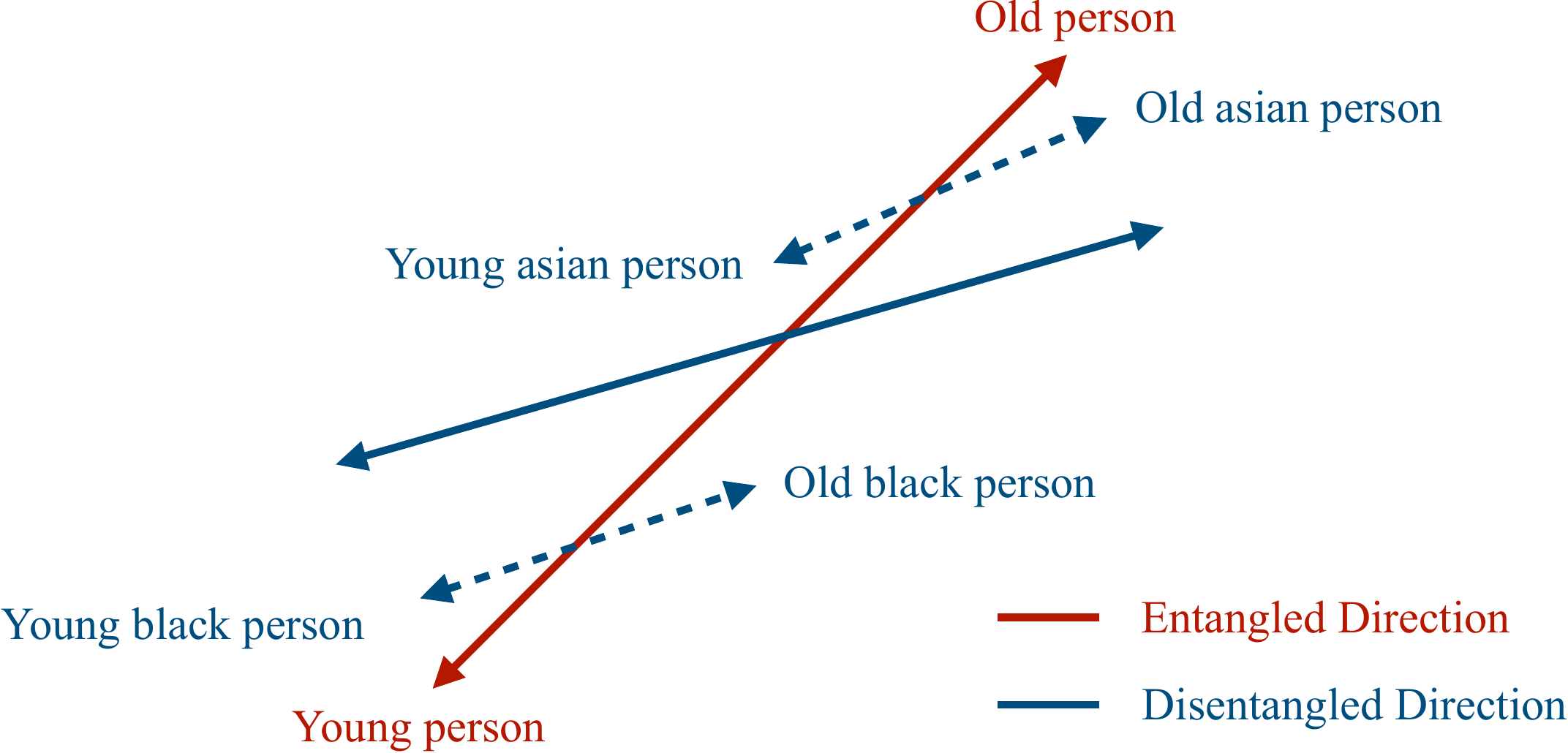}
    \caption{In this schematic we illustrate how multiple preservation concepts are used to disentangle a direction. For the sake of clarity in figure, we show examples for just two races. In practice, we preserve a diversity of several protected attribute directions.}
    \label{fig:directions}
\end{figure}

We visualize the rationale behind our disentangled formulation for sliders. When training sliders on single pair of prompts, sometimes the directions are entangled with unintended directions. For example, as we show show in Figure~\ref{fig:ablations}, controlling age can interfere with gender or race. We therefore propose using multiple paired prompts for finding a disentangled direction. As shown in Figure~\ref{fig:directions}, we explicitly define the preservation directions (dotted blue lines) to find a new edit direction (solid blue line) invariant to the preserve features.

\section{SDEdit Analysis}
\begin{figure}
    \centering
    \includegraphics[width=\linewidth]{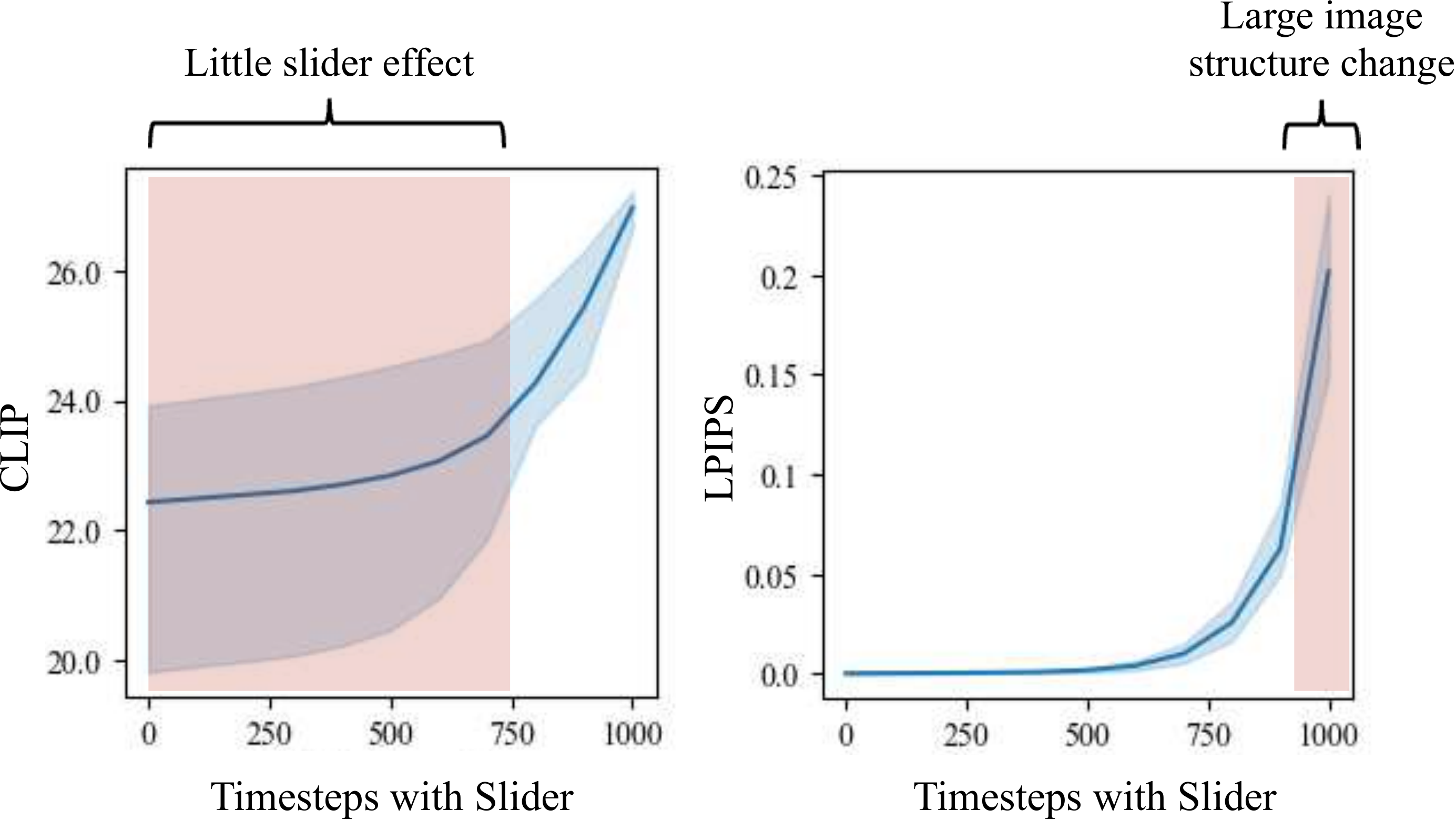}
    \caption{The plot examines CLIP score change and LPIPS distance when applying the same slider scale but with increasing SDEdit times. Higher timesteps enhance concept attributes considerably per CLIP while increased LPIPS demonstrates change in spatial stability. On the x-axis, 0 corresponds to no slider application while 1000 represents switching from start.}
    \label{fig:sdedit}
\end{figure}

We ablate SDEdit's contribution by fixing slider scale while varying SDEdit timesteps over 2,500 images. Figure \ref{fig:sdedit} shows inverse trends between LPIPS and CLIP distances as SDEdit time increases. Using more SDEdit maintains structure, evidenced by lower LPIPS score, while maintaining lower CLIP score. This enables larger slider scales before risking structural changes. We notice that on average, timestep 750 - 850 has the best of both worlds with spatial structure preservation and increased efficacy.

\section{Textual Concepts Sliders}
\begin{figure*}
    \centering
    \includegraphics[width=\linewidth]{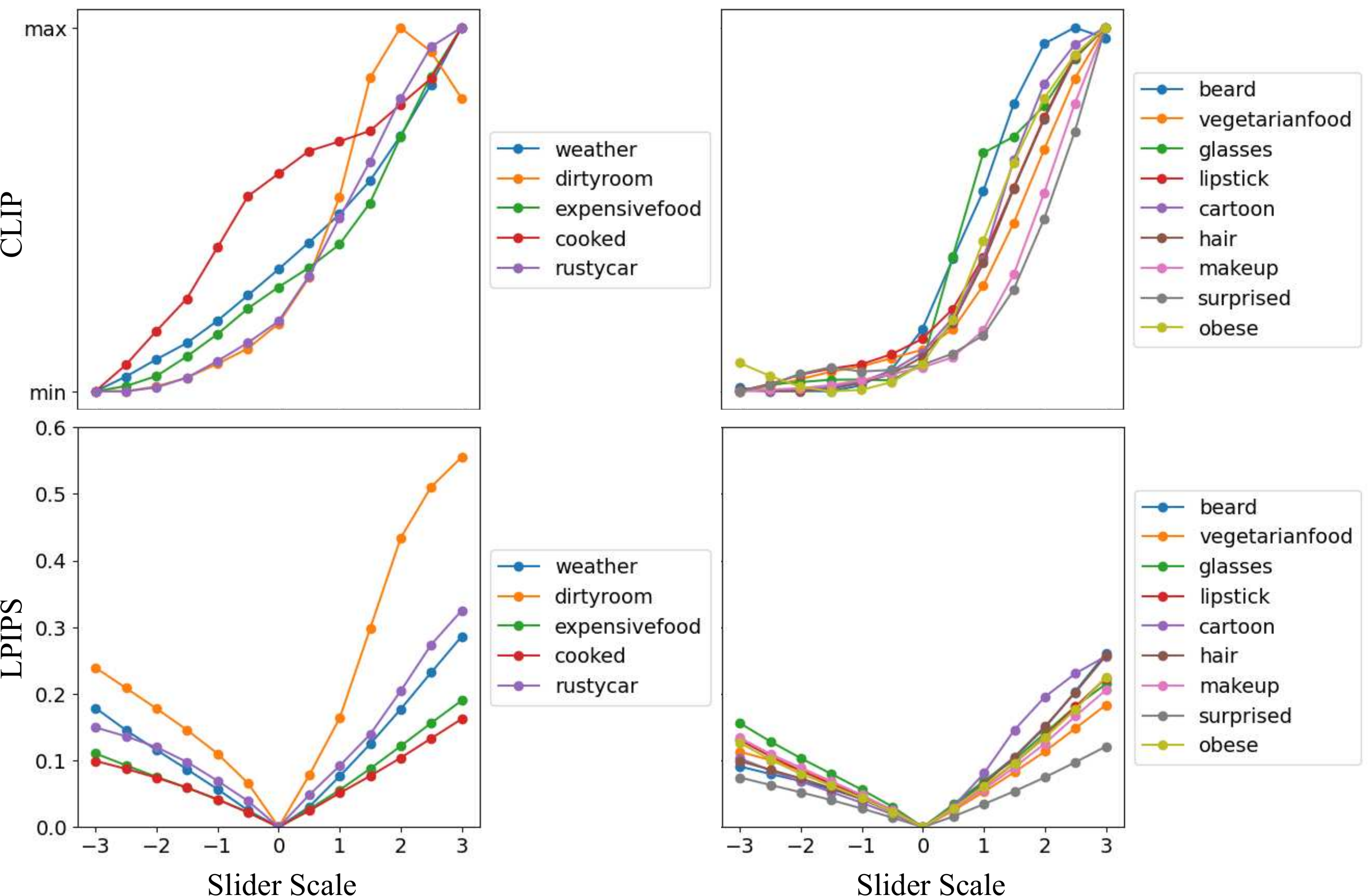}
    \caption{Analyzing attribute isolation efficacy vs stylistic variation for 15 slider types across 12 scales. We divide our figure into two columns. The left column contains concepts that have words for antonyms (\eg expensive - cheap) showing symmetric CLIP score deltas up/down. The right column shows harder to negate sliders (\eg no glasses) causing clipped negative range. We also note that certain sliders have higher lpips, such as ``cluttered'' room slider, which intuitively makes sense.}
    \label{fig:analyse_fine}
\end{figure*}

\begin{figure*}
    \centering
    \includegraphics[width=\linewidth]{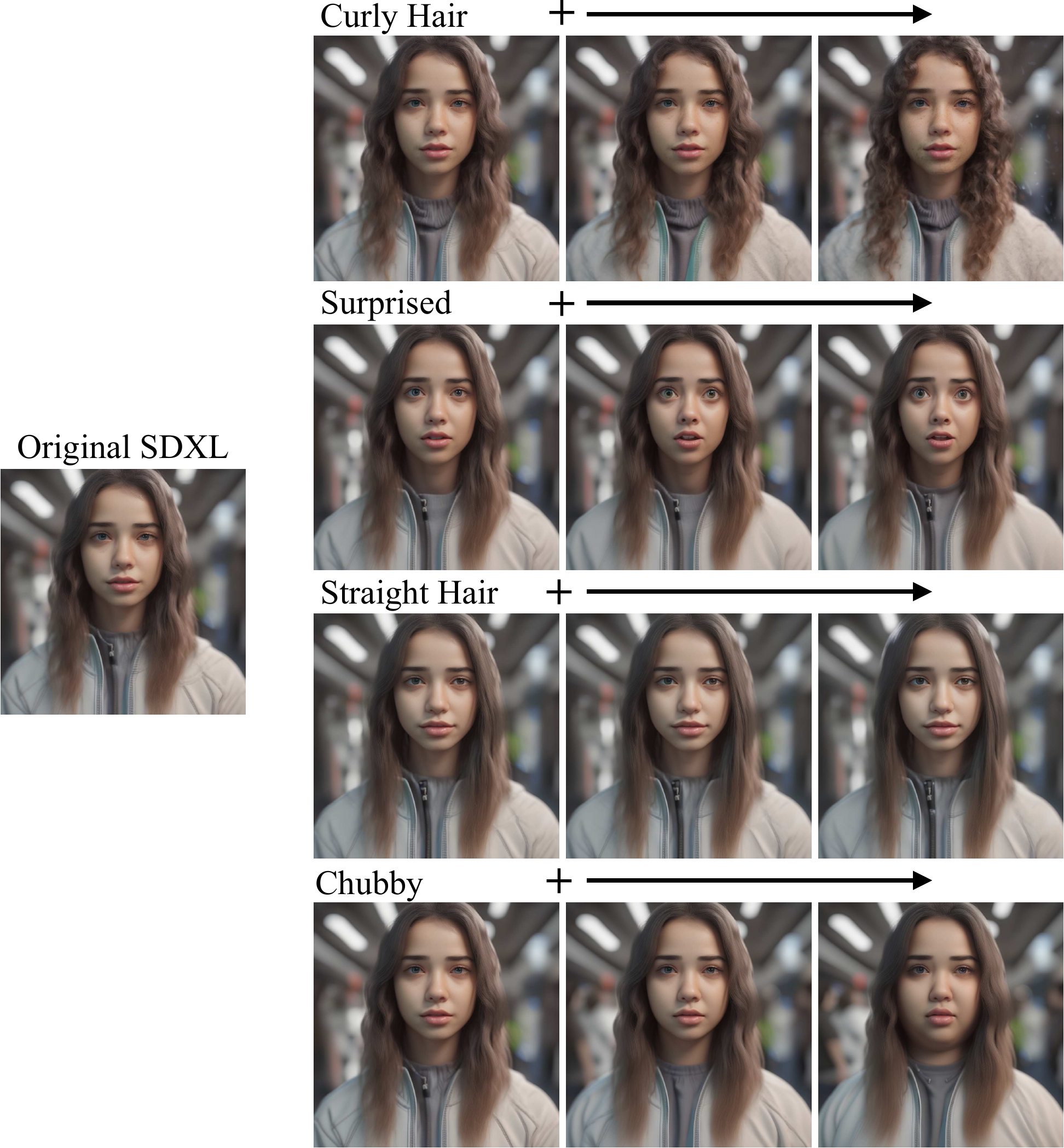}
    \caption{We demonstrate the effects of modifying an image with different sliders like ``curly hair'', ``surprised'', ``chubby''. Our text-based sliders allow precise editing of desired attributes during image generation while maintaining the overall structure.}
    \label{fig:text1}
\end{figure*}

\begin{figure*}
    \centering
    \includegraphics[width=\linewidth]{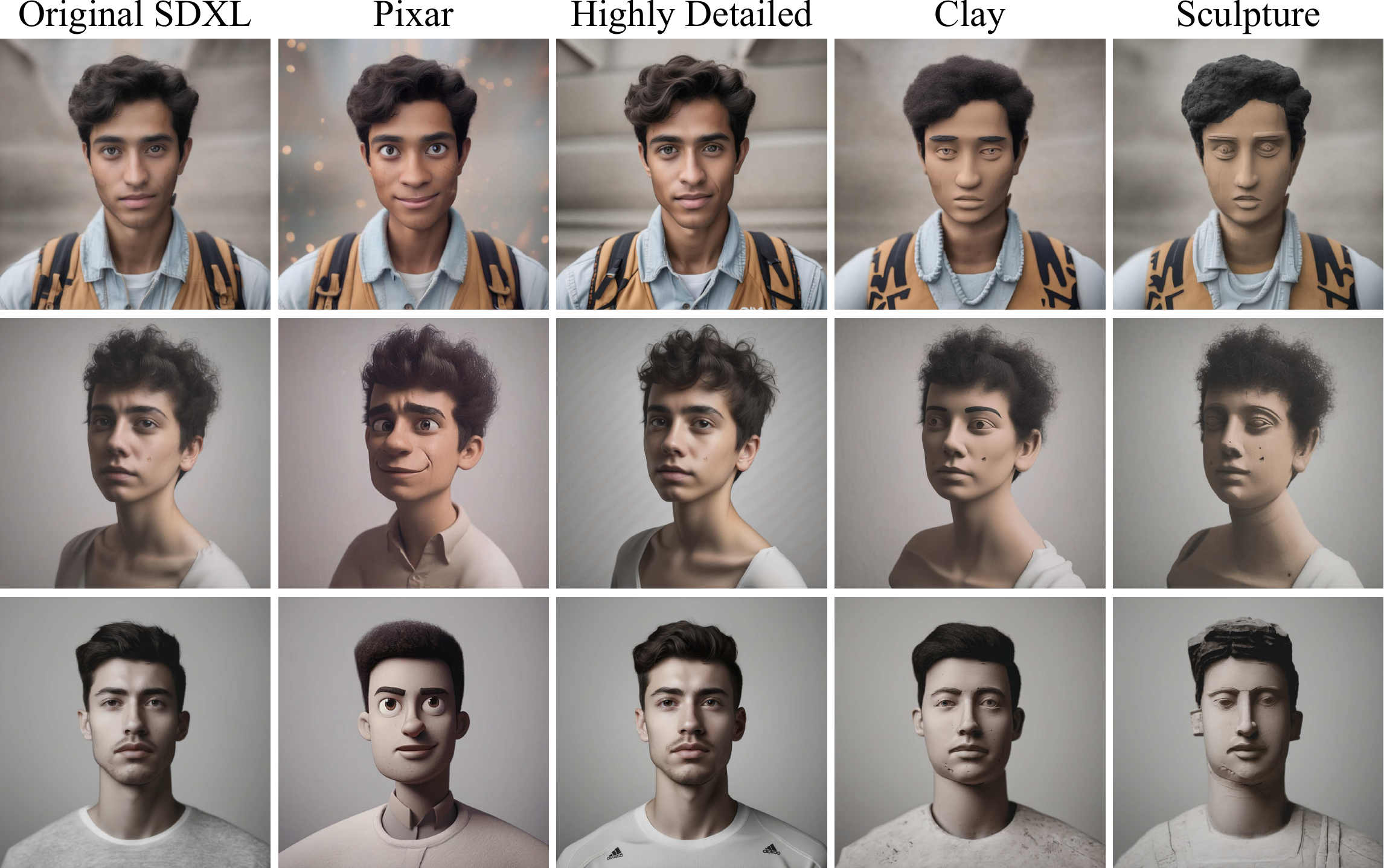}
    \caption{We demonstrate style sliders for "pixar", "realistic details", "clay", and "sculpture". Our text-based sliders allow precise editing of desired attributes during image generation while maintaining the overall structure.}
    \label{fig:text2}
\end{figure*}

\begin{figure*}
    \centering
    \includegraphics[width=\linewidth]{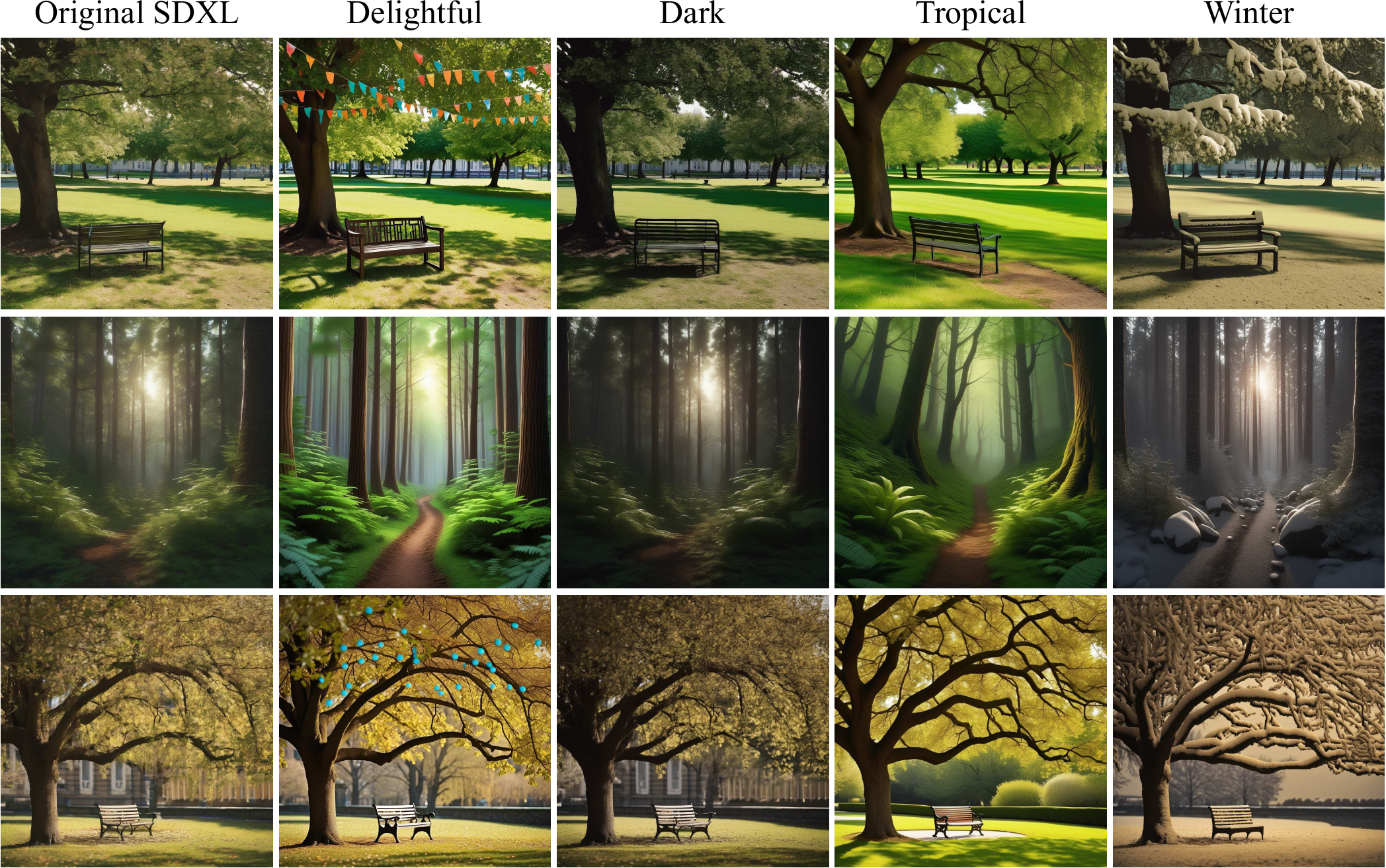}
    \caption{We demonstrate weather sliders for "delightful", "dark", "tropical", and "winter". For delightful, we notice that the model sometimes make the weather bright or adds festive decorations. For tropical, it adds tropical plants and trees. Finally, for winter, it adds snow.}
    \label{fig:text3}
\end{figure*}

\begin{figure*}
    \centering
    \includegraphics[width=\linewidth]{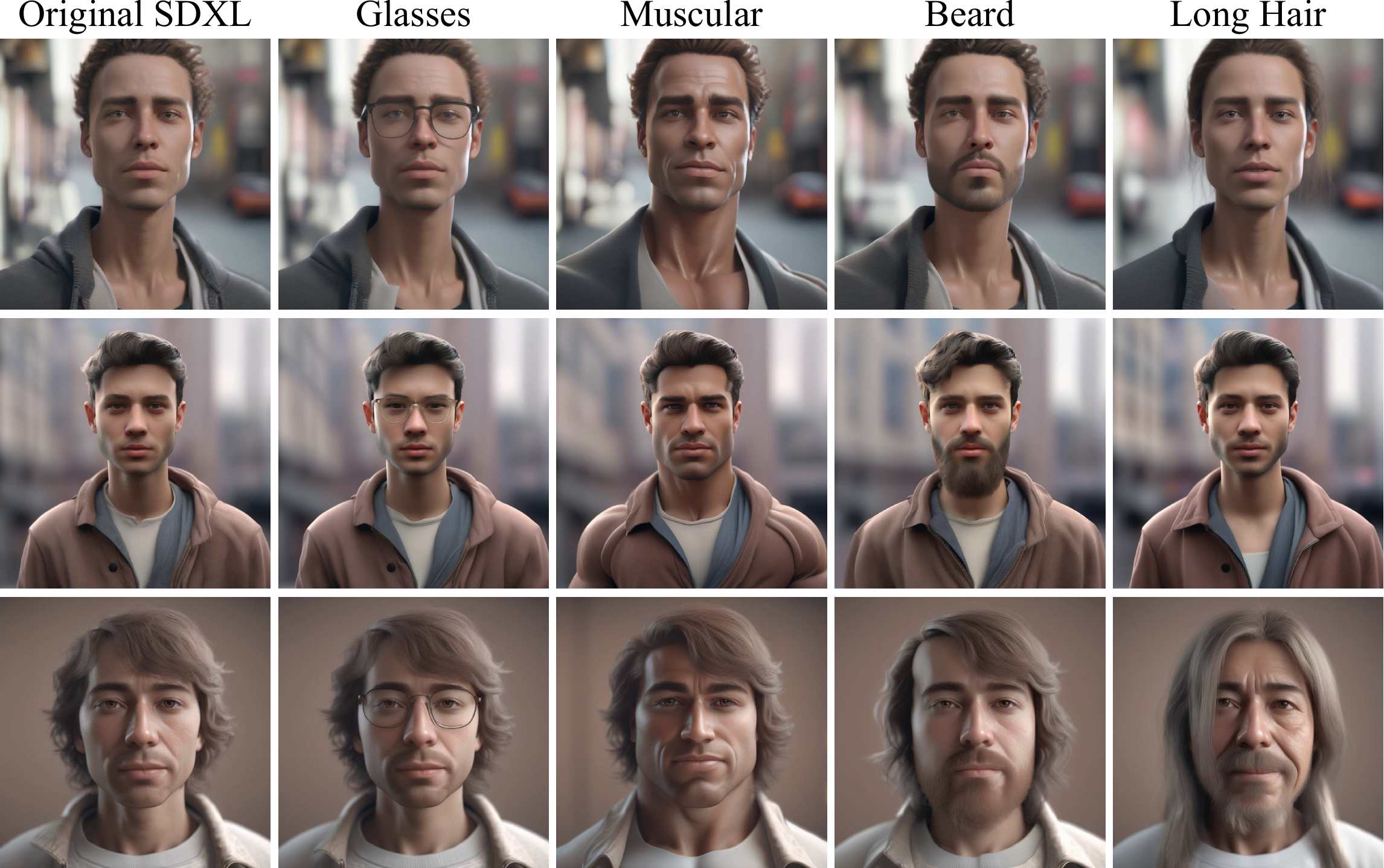}
    \caption{We demonstrate sliders to add attributes to people like "glasses", "muscles", "beard", and "long hair". Our text-based sliders allow precise editing of desired attributes during image generation while maintaining the overall structure.}
    \label{fig:text4}
\end{figure*}

\begin{figure*}
    \centering
    \includegraphics[width=\linewidth]{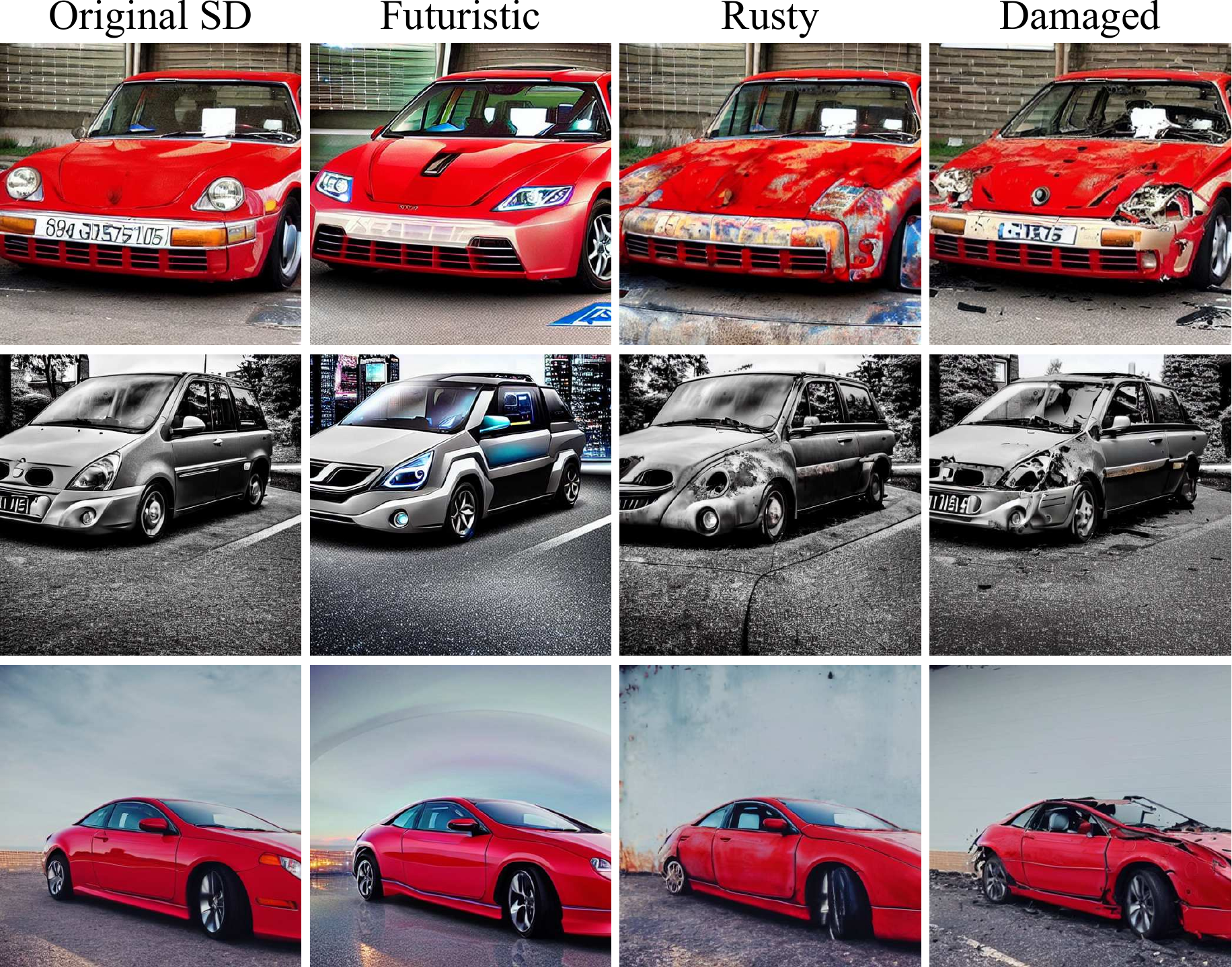}
    \caption{We demonstrate sliders to control attributes of vehicles like ``rusty'', ``futuristic'', ``damaged''. Our text-based sliders allow precise editing of desired attributes during image generation while maintaining the overall structure.}
    \label{fig:text5}
\end{figure*}

\begin{figure*}
    \centering
    \includegraphics[width=\linewidth]{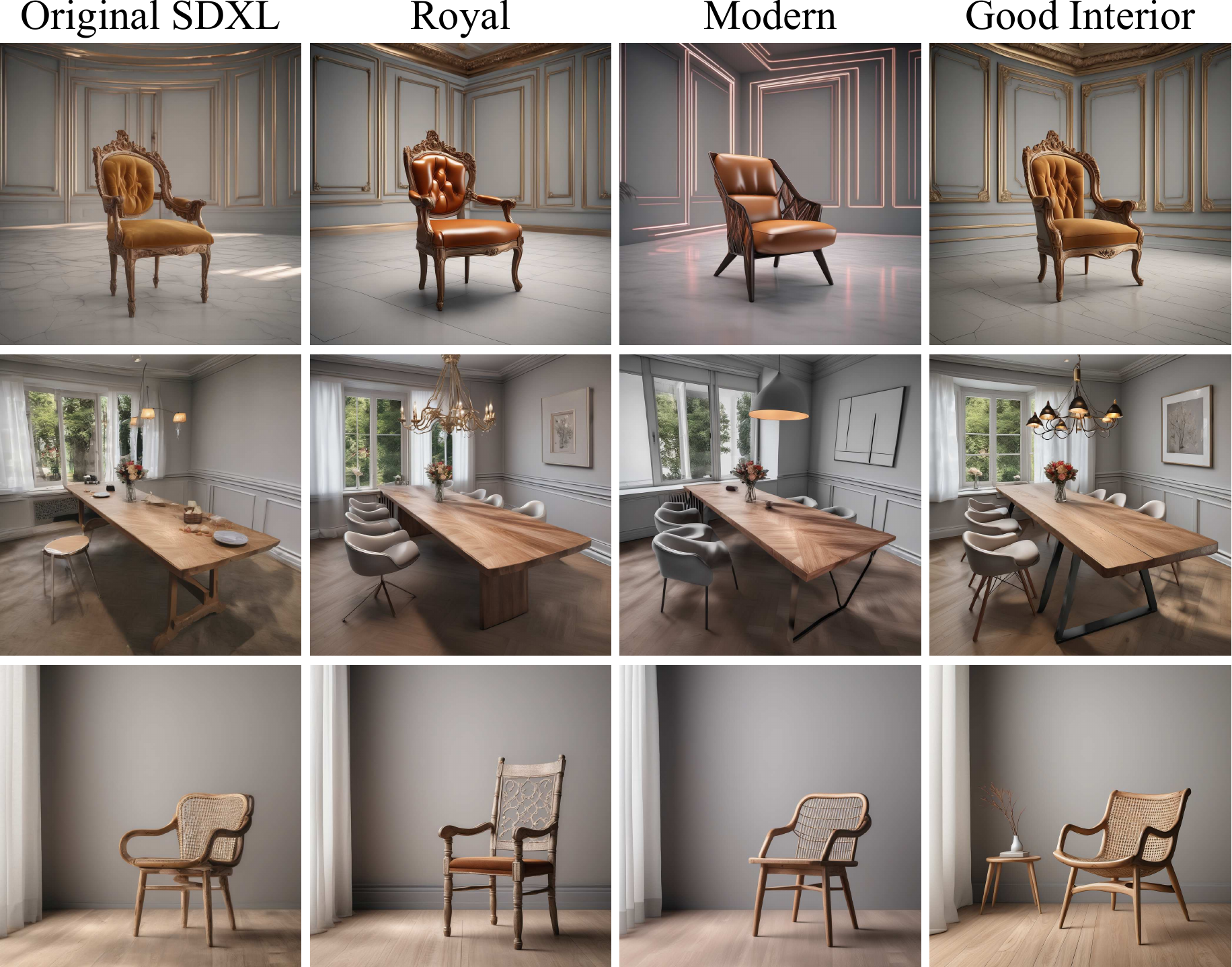}
    \caption{Our sliders can also be used to control styles of furniture like ``royal'', ``Modern''. Our text-based sliders allow precise editing of desired attributes during image generation while maintaining the overall structure.}
    \label{fig:text6}
\end{figure*}

We quantify slider efficacy and control via CLIP score change and LPIPS distance over 15 sliders at 12 scales in Figure~\ref{fig:analyse_fine}. CLIP score change validates concept modification strength. Tighter LPIPS distributions demonstrate precise spatial manipulation without distortion across scales. We show additional qualitative examples for textual concept sliders in Figures~\ref{fig:text1}-\ref{fig:text6}.
\subsection{Baseline Details}
We compare our method against Prompt-to-prompt and a novel inference-time prompt composition method. For Prompt-to-prompt we use the official implementation code\footnote{\href{https://github.com/google/prompt-to-prompt/}{https://github.com/google/prompt-to-prompt/}}. We use the Refinement strategy they propose, where new token is added to the existing prompt for image editing. For example, for the images in Figure~\ref{fig:baseline_entangle}, we add the token ``old'' for the original prompt ``picture of person'' to make it ``picture of old person''. For the compostion method, we use the principles from Liu et al~\footnote{\href{https://energy-based-model.github.io/Compositional-Visual-Generation-with-Composable-Diffusion-Models/}{https://energy-based-model.github.io/Compositional-Visual-Generation-with-Composable-Diffusion-Models/}}. Specifically, we compose the score functions coming from both ``picture of person'' and ``old person'' through additive guidance. We also utilize the SDEdit technique for this method to allow finer image editing.

\subsection{Entanglement}
\begin{figure}
    \centering
    \includegraphics[width=\linewidth]{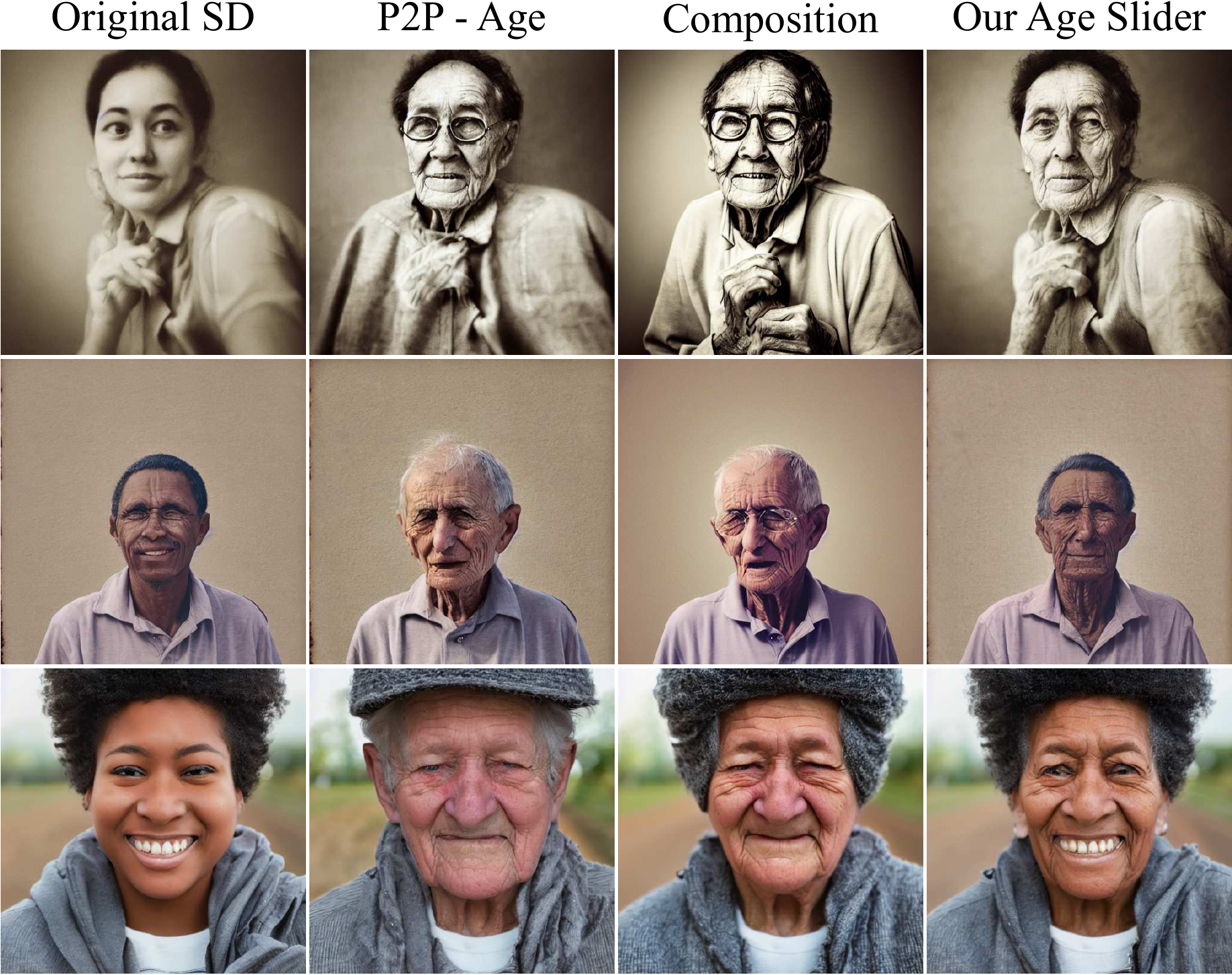}
    \caption{Concept Sliders demonstrate minimal entanglement when controlling a concept. Prompt-to-prompt and inference-time textual composition sometimes tend to alter race/gender when editing age.}
    \label{fig:baseline_entangle}
\end{figure}

The baselines are sometimes prone to interference with concepts when editing a particular concept. Table~\ref{tab:baseline_entangle} shows quantitative analysis on interference while Figure~\ref{fig:baseline_entangle} shows some qualititative examples. We find that Prompt-to-prompt and inference composition can sometimes change the race/gender when editing age. Our sliders with disentaglement object~\ref{eq:objective_disentangle}, show minimal interference as seen by \textit{Interference} metric, which shows the percentage samples with race or gender changed out of 2500 images we tested. We also found through LPIPS metric that our method shows finer editing capabilities. We find similar conclusions through quanlitative samples from Figure~\ref{fig:baseline_entangle}, that P2P and composition can alter gender, race or both when controlling age. 

\begin{table}[ht]
    \centering
    \begin{tabular}{rccc}
    &  \textbf{P2P}& \textbf{Composition} &\textbf{Ours}\\
    \hline
    $\mathbf{\Delta_{CLIP}}$  & 1.10 &3.14 & \textbf{3.93}  \\ 
    \textbf{LPIPS}          &0.15&0.13 &\textbf{0.06}\\
    \textbf{Interference}    &0.33&0.38 &\textbf{0.10}\\
    \end{tabular}
    \caption{The disentanglement formulation enables precise control over the age direction, as shown by the significant reduction in the Interference metric which measures the percentage of samples with gender/race change, compared to the original images. By using LoRA adaptors, sliders achieve finer editing in terms of both structure and edit direction, as evidenced by improvements in LPIPS and Interference. Concept strength is maintained, with similar $\Delta_{CLIP}$ scores across ablations.}
    \label{tab:baseline_entangle}
\end{table}

\section{Visual Concept}
\subsection{Baseline Details}
We compare our method to two image customization baselines: custom diffusion~\footnote{\href{https://github.com/adobe-research/custom-diffusion}{https://github.com/adobe-research/custom-diffusion}} and textual inversion~\footnote{\href{https://github.com/rinongal/textual_inversion}{https://github.com/rinongal/textual\_inversion}}. For fair comparison, we use the official implementations of both, modifying textual inversion to support SDXL. These baselines learn concepts from concept-labeled image sets. However, this approach risks entangling concepts with irrelevant attributes (e.g. hair, skin tone) that correlate spuriously in the dataset, limiting diversity.
\subsection{Precise Concept Capturing}
\begin{figure}
   \centering
   \includegraphics[width=1\linewidth]{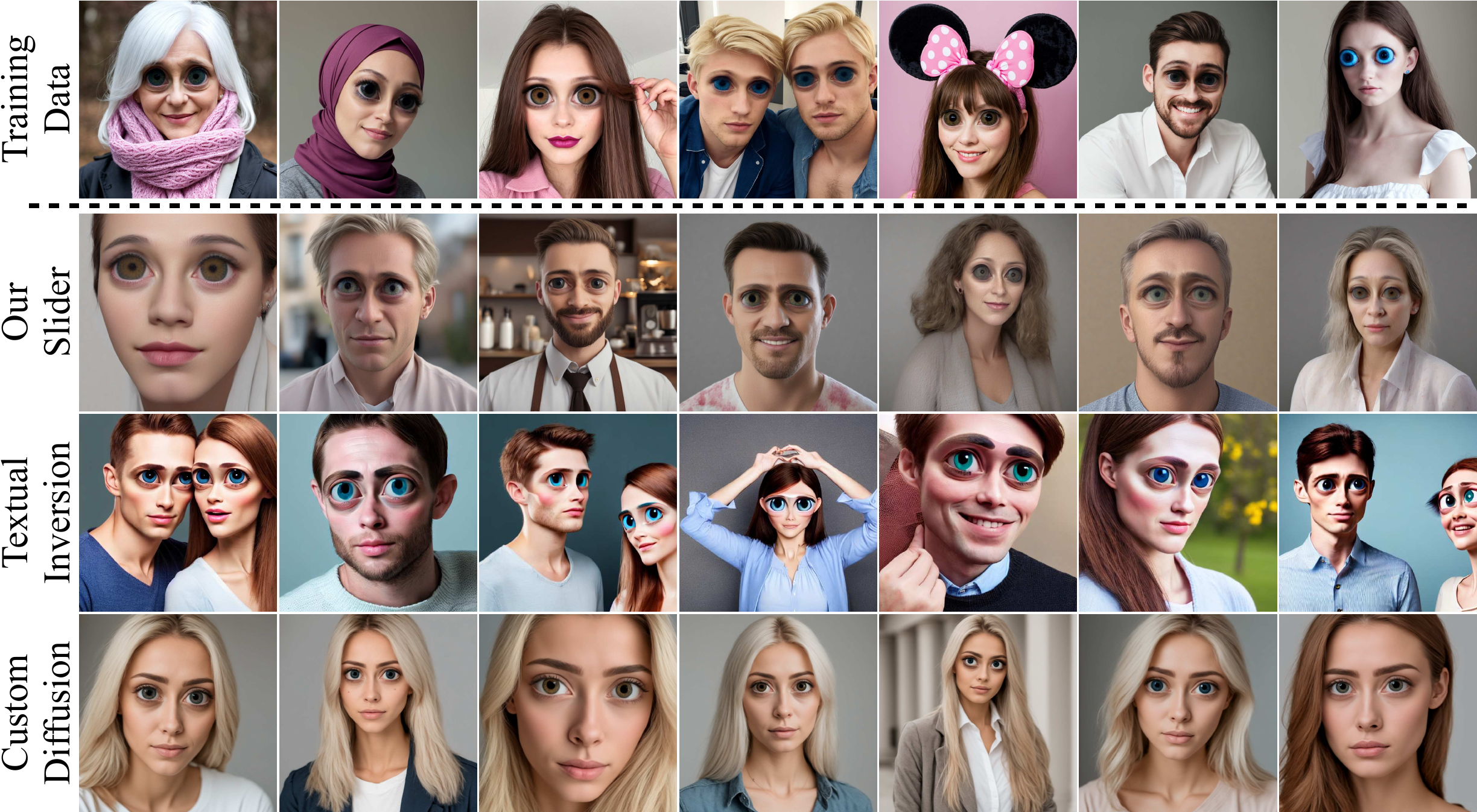}
   \caption{Concept Sliders demonstrate more diverse outputs while also being effective at learning the new concepts. Customization methods can sometimes tend to learn unintended concepts like hair and eye colors.}
   \label{fig:visual_diverse}
\end{figure}

Figure \ref{fig:visual_diverse} shows non-cherry-picked customization samples from all methods trained on the large-eyes Ostris dataset~\footnote{\href{https://github.com/ostris/ai-toolkit}{https://github.com/ostris/ai-toolkit}}. While exhibiting some diversity, samples frequently include irrelevant attributes correlated with large eyes in the dataset, \eg blonde hair in custom diffusion, blue eyes in textual inversion. In contrast, our paired image training isolates concepts by exposing only local attribute changes, avoiding spurious correlation learning.

\section{Composing Sliders}
\begin{figure}
    \centering
    \includegraphics[width=\linewidth]{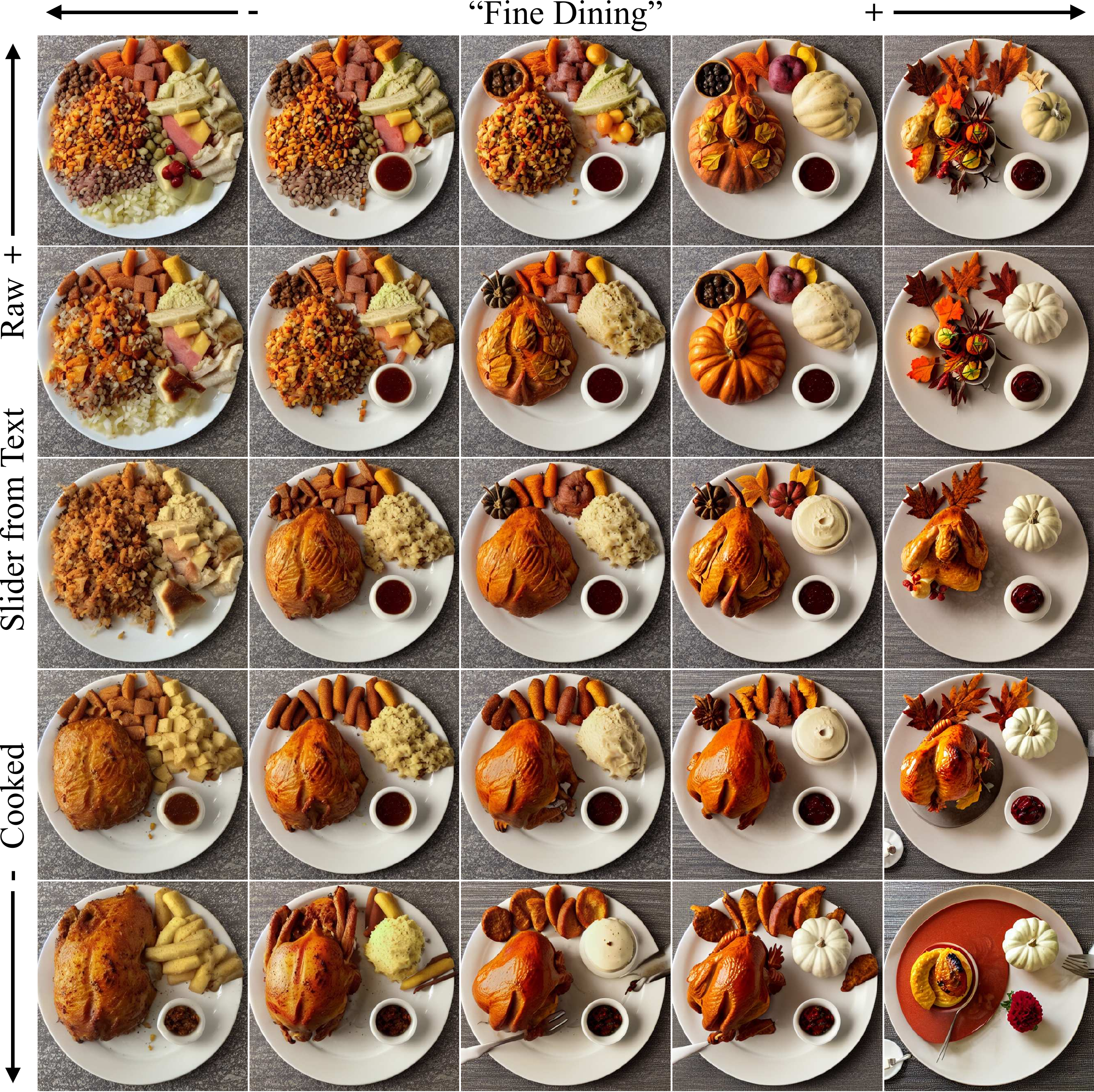}
    \caption{Composing two text-based sliders results in a complex control over thanksgiving food options. We show the effect of applying both the "cooked" slider and "fine-dining" slider to a generated image of thanksgiving dinner. These sliders can be used in both positive and negative directions.}
    \label{fig:2d_food_thanks}
\end{figure}

\begin{figure}
    \centering
    \includegraphics[width=\linewidth]{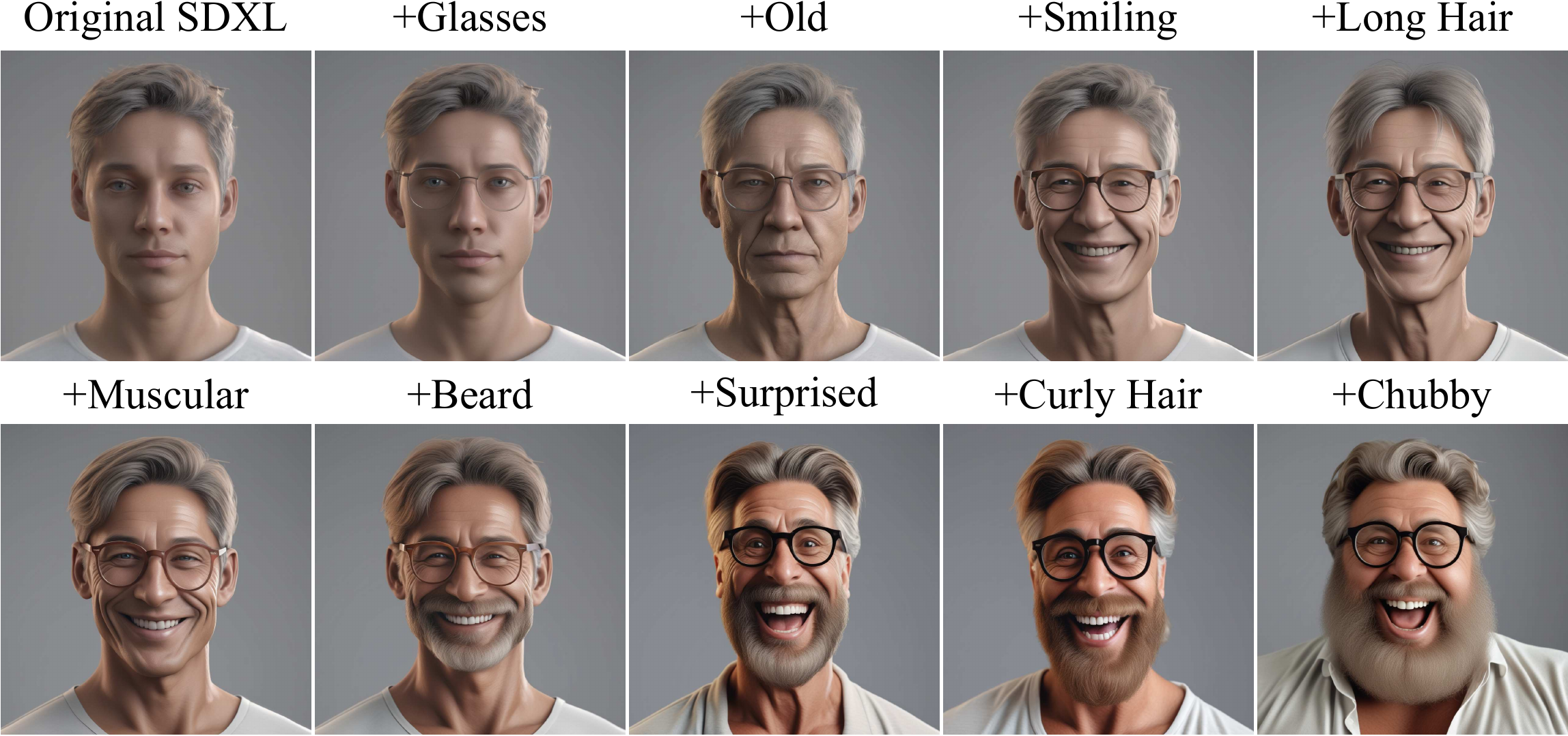}
    \caption{Concept Sliders can be composed for a more nuanced and complex control over attributes in an image. From stable diffusion XL image on the top left, we progressively compose a slider on top of the previously added stack of sliders. By the end, bottom right, we show the image by composing all 10 sliders.}
    \label{fig:compose1}
\end{figure}

\begin{figure}
    \centering
    \includegraphics[width=\linewidth]{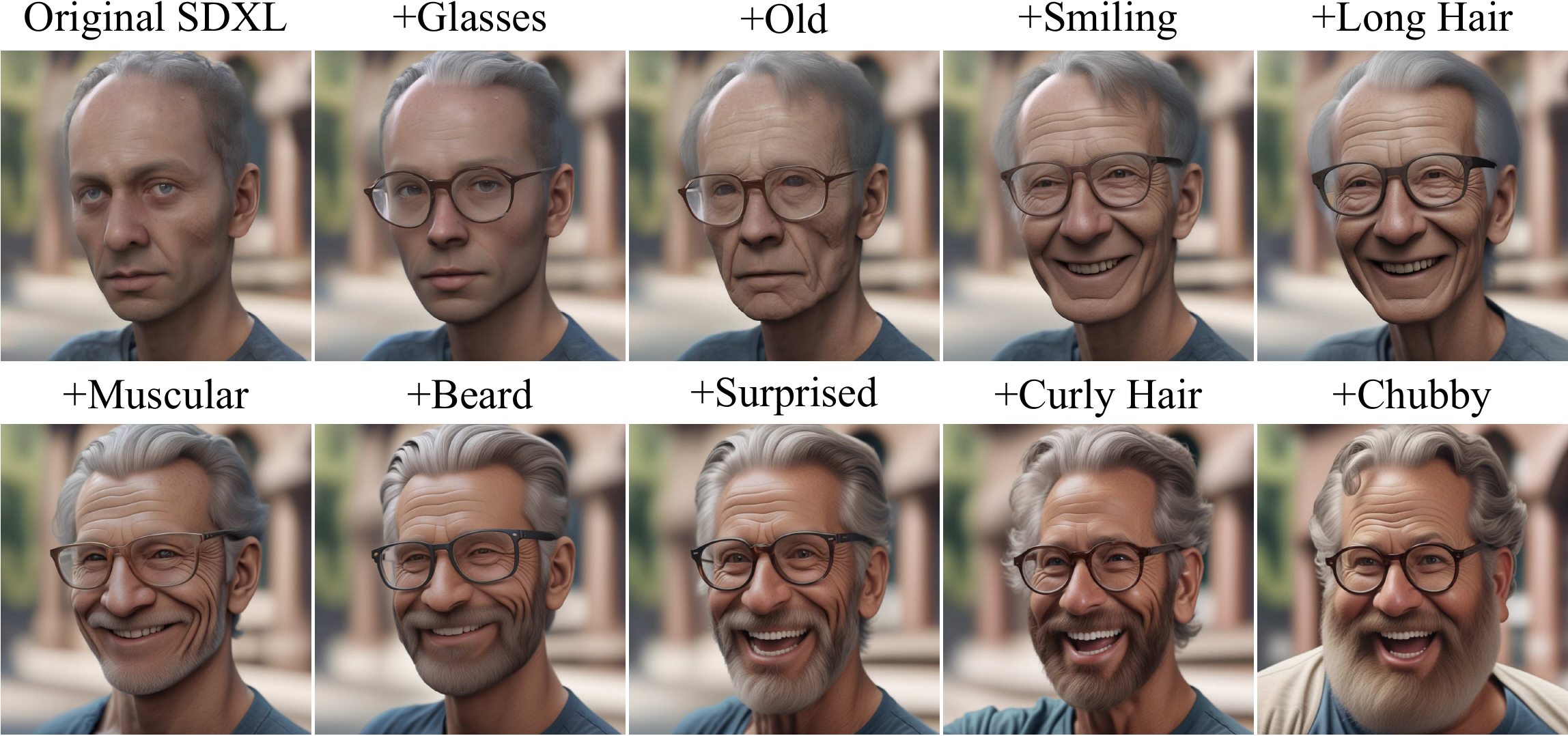}
    \caption{Concept Sliders can be composed for a more nuanced and complex control over attributes in an image. From stable diffusion XL image on the top left, we progressively compose a slider on top of the previously added stack of sliders. By the end, bottom right, we show the image by composing all 10 sliders.}
    \label{fig:compose2}
\end{figure}

We show a 2 dimensional slider by composing ``cooked'' and ``fine dining'' food sliders in Figure~\ref{fig:2d_food_thanks}. Next, we show progessive composition of sliders one by one in Figures~\ref{fig:compose1},\ref{fig:compose2}. From top left image (original SDXL), we progressively generate images by composing a slider at each step. We show how our sliders provide a semantic control over images. 


\section{Editing Real Images}
\begin{figure}
   \centering
   \includegraphics[width=1\linewidth]{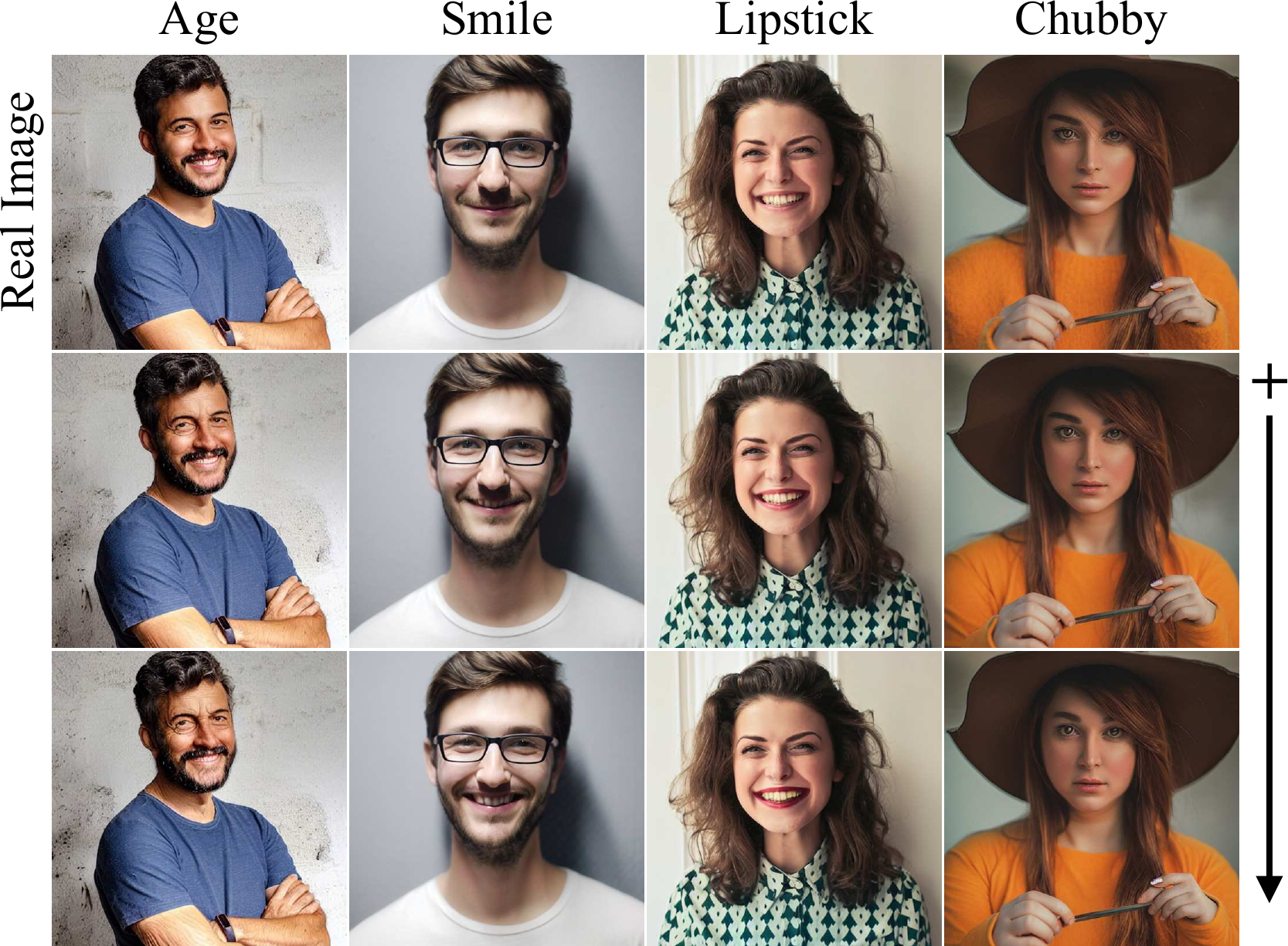}
   \caption{Concept Sliders can be used to edit real images. We use null inversion method to convert real image as a diffusion model generated image. We then run our Concept Sliders on that generation to enable precise control of concepts. }
   \label{fig:realimage}
\end{figure}

Concept sliders can also be used  to edit real images. Manually engineering a prompt to generate an image similar to the real image is very difficult. We use null inversion~\footnote{\href{https://null-text-inversion.github.io}{https://null-text-inversion.github.io}} which finetunes the unconditional text embedding in the classifier free guidance during inference. This allows us to find the right setup to turn the real image as a diffusion model generated image. Figure~\ref{fig:realimage} shows Concept Sliders used on real images to precisely control attributes in them.

\section{Sliders to Improve Image Quality}
\begin{figure}
    \centering
    \includegraphics[width=\linewidth]{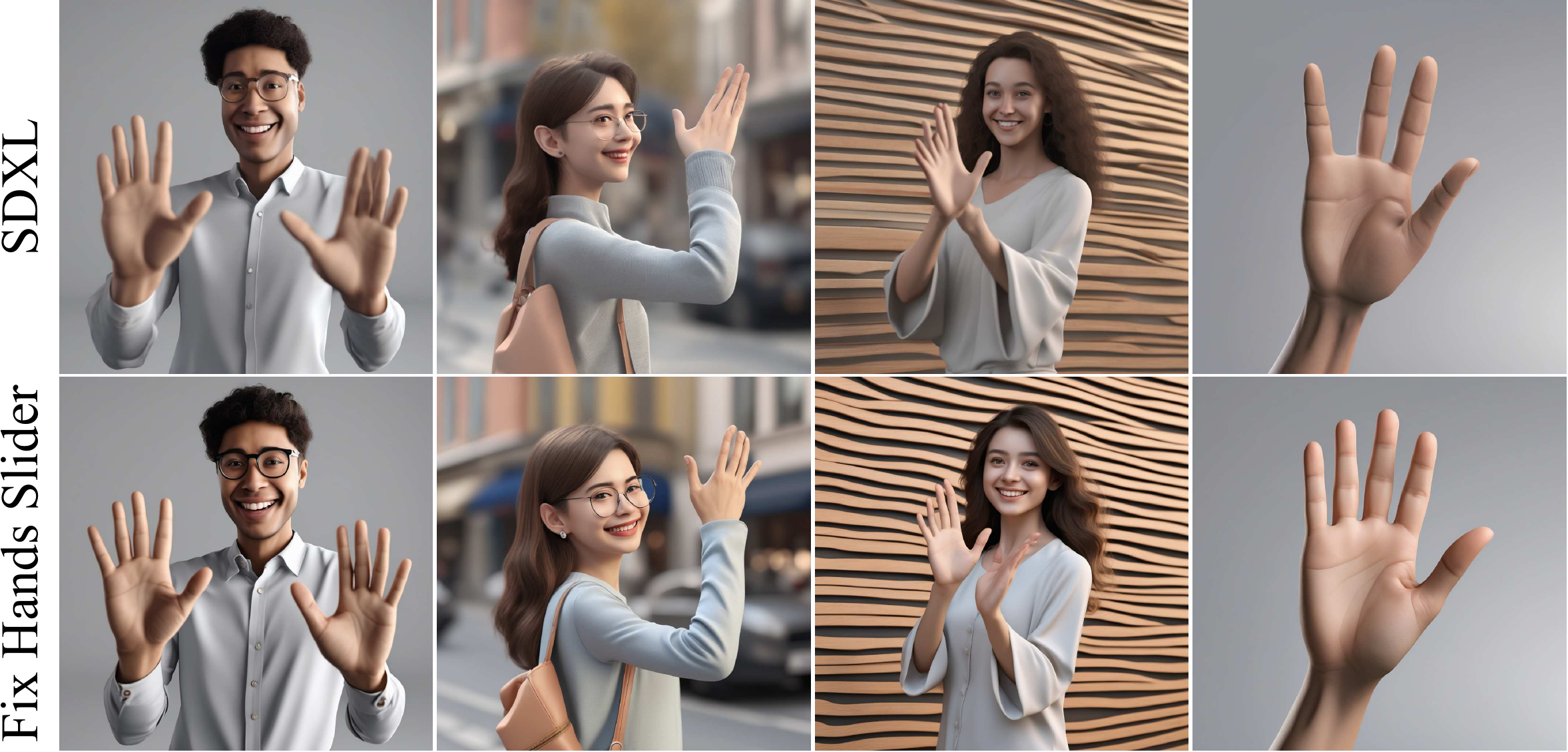}
    \caption{Concept Sliders can be used to fix common distortions in diffusion model generated images. We demonstrate "Fix Hands" slider that can fix distorted hands.}
    \label{fig:hands1}
\end{figure}

\begin{figure*}
    \centering
    \includegraphics[width=\linewidth]{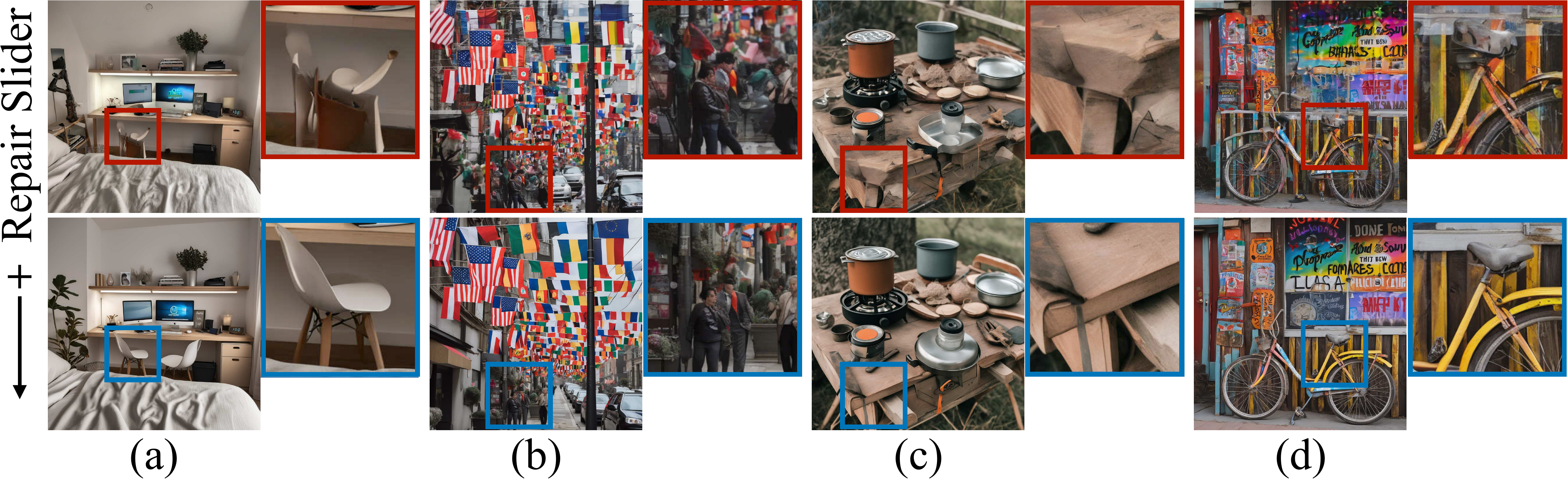}
    \caption{Concept Sliders can be used to fix common distortions in diffusion model generated images. The repair slider enables the model to generate images that are more realistic and undistorted.}
    \label{fig:realistic1}
\end{figure*}

\begin{figure*}
    \centering
    \includegraphics[width=\linewidth]{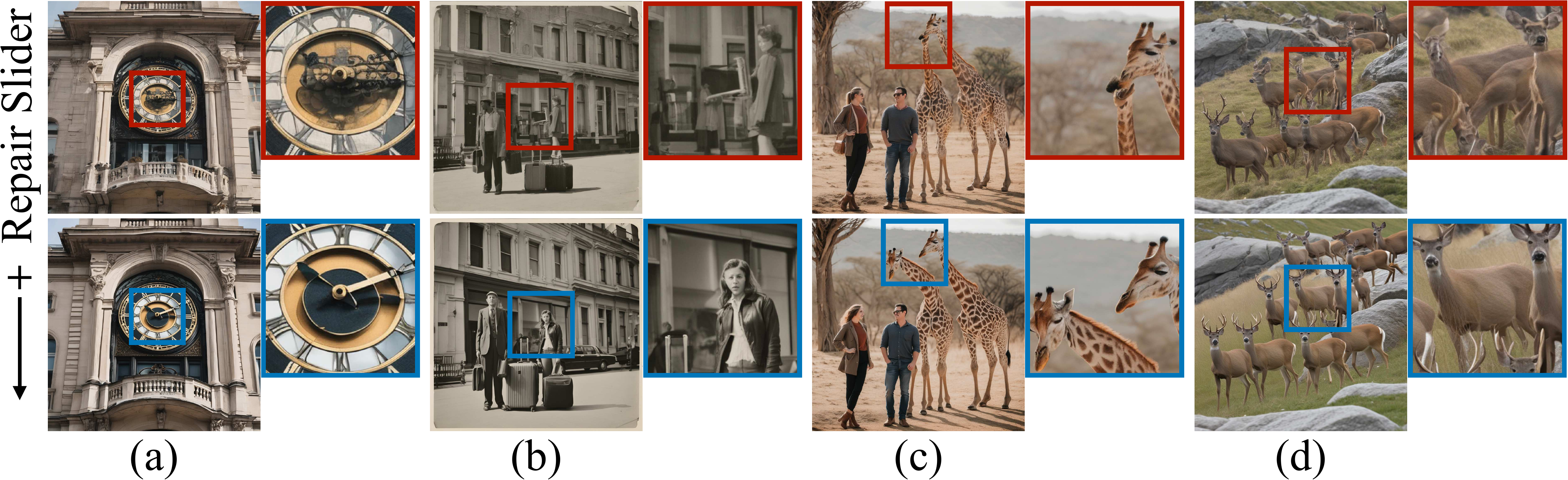}
    \caption{Concept Sliders can be used to fix common distortions in diffusion model generated images. The repair slider enables the model to generate images that are more realistic and undistorted.}
    \label{fig:realistic2}
\end{figure*}

\begin{figure*}
    \centering
    \includegraphics[width=\linewidth]{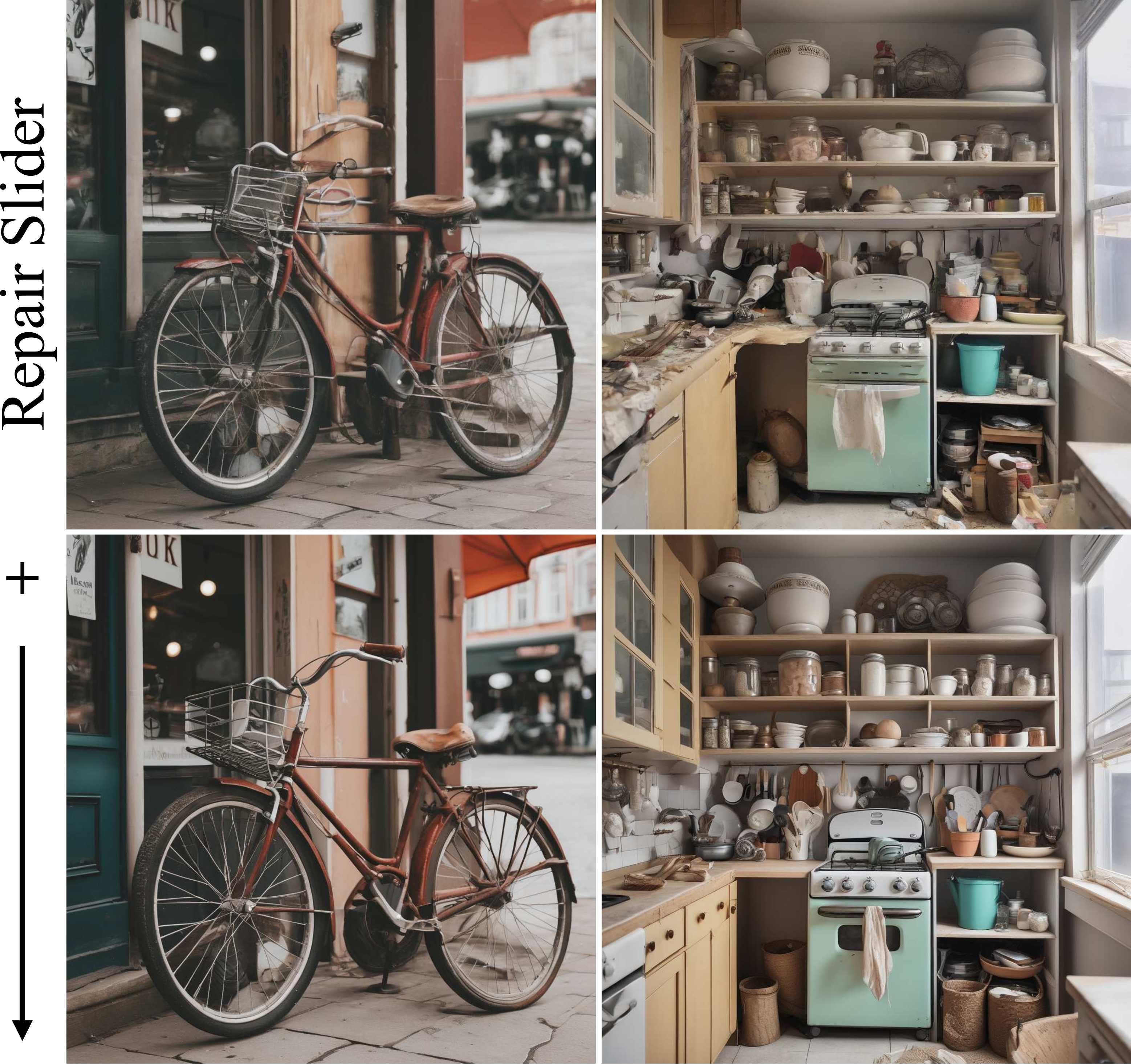}
    \caption{Concept Sliders can be used to fix common distortions in diffusion model generated images. The repair slider enables the model to generate images that are more realistic and undistorted.}
    \label{fig:realistic3}
\end{figure*}

We provide more qualitative examples for "fix hands" slider in Figure~\ref{fig:hands1}. We also show additional examples for the "repair" slider in Figure~\ref{fig:realistic1}-\ref{fig:realistic3}

\subsection{Details about User Studies}
\begin{figure*}
    \centering
    \includegraphics[width=\linewidth]{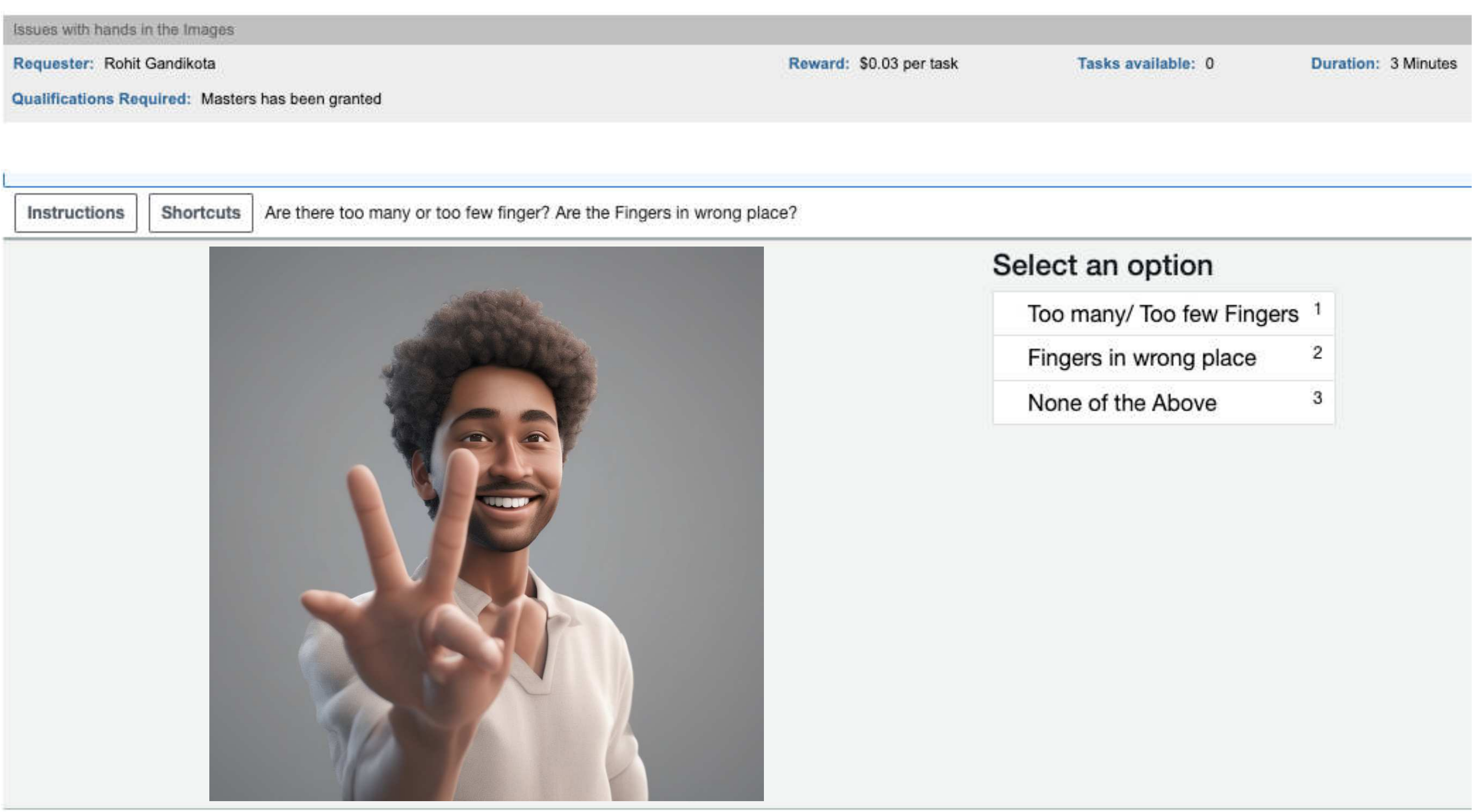}
    \caption{User study interface on Amazon Mechanical Turk. Users are shown images randomly sampled from either SDXL or our ``fix hands'' slider method, and asked to identify hand issues or mark the image as free of errors. Aggregate ratings validate localization capability of our finger control sliders. For the example shown above, users chose the option ``Fingers in wrong place''}
    \label{fig:user_hands}
\end{figure*}

We conduct two human evaluations analyzing our ``repair'' and ``fix hands'' sliders. For ``fix hands'', we generate 150 images each from SDXL and our slider using matched seeds and prompts. We randomly show each image to an odd number users and have them select issues with the hands: 1) misplaced/distorted fingers, 2) incorrect number of fingers, 3) none. as shown in Figure~\ref{fig:user_hands}
62\% of the 150 SDXL images have hand issues as rated by a majority of users. In contrast, only 22\% of our method's images have hand issues, validating effectiveness of our fine-grained control.\par

\begin{figure*}
    \centering
    \includegraphics[width=\linewidth]{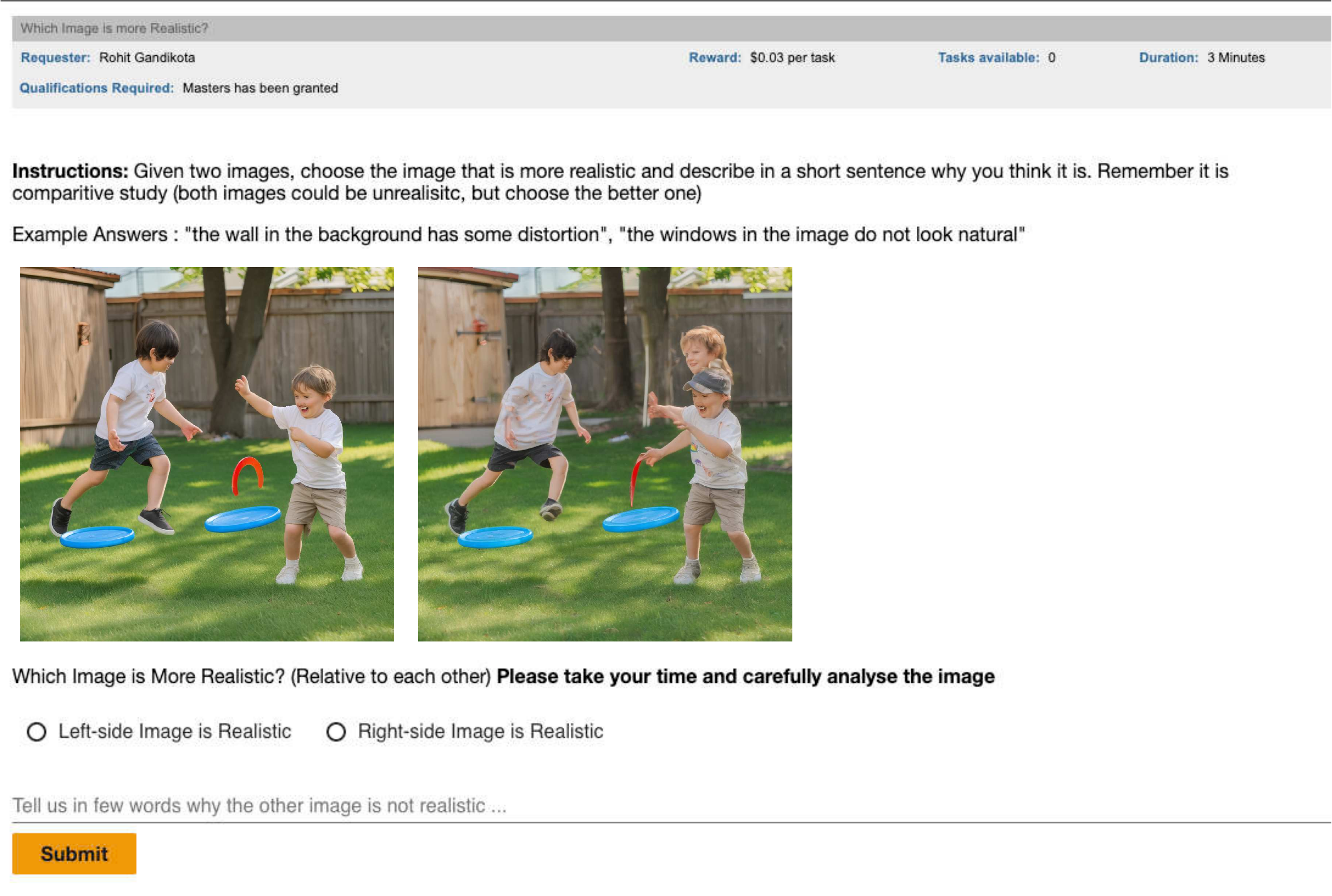}
    \caption{ Interface for our "realistic" slider user study. Users are shown an original SDXL image and the corresponding output from our slider, with left/right placement randomized. Users select the image they find more photorealistic and describe their rationale as a sanity check. For example, one user selected the slider image as more realistic in the shown example, commenting ``\textit{The black-haired boy's face, right arm and left foot are distorted in right image.}'' Another user also chose the slider output, noting ``\textit{The right side image has a floating head}''. Asking raters to give reasons aims to reduce random selections.}
    \label{fig:user_real}
\end{figure*}

We conduct an A/B test to evaluate the efficacy of our proposed ``repair''it slider. The test set consists of 300 image pairs (Fig. \ref{fig:user_real}), where each pair contains an original image alongside the output of our method when applied to that image with the same random seed. The left/right placement of these two images is randomized. Through an online user study, we task raters to select the image in each pair that exhibits fewer flaws or distortions, and to describe the reasoning behind their choice as a sanity check. For example, one rater selected the original image in Fig. \ref{fig:realistic1}.a, commenting that ``\textit{The left side image is not realistic because the chair is distorted.}'' . Similarly a user commented ``\textit{Giraffes heads are separate unlikely in other image}'' for Fig.~\ref{fig:realistic2}.c. Across all 300 pairs, our ``repair'' slider output is preferred as having fewer artifacts by 80.39\% of raters. This demonstrates that the slider effectively reduces defects relative to the original. We manually filter out responses with generic comments (e.g., ``more realistic''), as the sanity check prompts raters for specific reasons. After this filtering, 250 pairs remain for analysis.
 
\end{document}